\documentclass[letterpaper,journal]{IEEEtran}
\usepackage{amsmath,amsfonts}
\usepackage{algorithmic}
\usepackage{algorithm}
\usepackage{array}
\usepackage[caption=false,font=normalsize,labelfont=sf,textfont=sf]{subfig}
\usepackage{textcomp}
\usepackage{stfloats}
\usepackage{url}
\usepackage{verbatim}
\usepackage{graphicx}
\usepackage{tikz}

\usepackage{tikz}
\usetikzlibrary{arrows.meta,positioning,calc,fit}
\tikzset{
  tkdebox/.style={
    draw=black!60,line width=0.45pt,rounded corners=1.5pt,
    fill=white,align=left,inner xsep=6pt,inner ysep=5pt,outer sep=0pt
  },
  tkdearrow/.style={-{Stealth[length=1.6mm,width=1.1mm]},
    draw=black!70,line width=0.65pt},
  tkdefeedback/.style={tkdearrow,dashed},
  tkdelabel/.style={font=\fontsize{8}{9.5}\selectfont,
    align=center,inner sep=1pt,fill=white,text=black!80}
}

\usepackage{tabularx}
\usepackage{booktabs}
\usepackage{amssymb} 
\usepackage{cite}
\definecolor{tbBlue}{HTML}{4875A5}
\definecolor{tbRust}{HTML}{996747}
\definecolor{tbGreen}{HTML}{388675}
\definecolor{tbInk}{HTML}{23313D}
\definecolor{tbGray}{HTML}{66717B}
\definecolor{tbHead}{HTML}{EDF1F4}
\definecolor{tbBlueFill}{HTML}{E7EFF6}
\definecolor{tbPeachFill}{HTML}{FBEEE7}
\definecolor{tbGreenFill}{HTML}{E6F1E7}

\newcommand{\TableHeader}[1]{%
  \colorbox{tbHead}{%
    \parbox[t][2.2\baselineskip][c]
      {\dimexpr\linewidth-2\fboxsep\relax}{%
        \raggedright\bfseries #1}}}

\newcommand{\TableBand}[3]{%
  \colorbox{#1}{%
    \parbox{\dimexpr\textwidth-2\fboxsep\relax}{%
      \strut\textcolor{#2}{\bfseries #3}\strut}}}

\begin{document}

\bstctlcite{terminal_ieee_control}

\title{Graph-Mediated Large Language Model Systems in the Agentic Era: Reasoning, Knowledge, Retrieval, Execution, and Coordination}

\author{Haihui Liu, Yang Chen, Zice Qiu, Yun Zhang, Jialun Zhong, Wenhua Wang,~\IEEEmembership{Member,~IEEE}, Yanzeng Li,~\IEEEmembership{Member,~IEEE} and Tian Wang,~\IEEEmembership{Senior Member,~IEEE}

\thanks{The work was supported in part by the Joint Funds of the National Natural Science Foundation of China under Grant U25A20436, the National Natural Science Foundation of China (NSFC) under Grants 62372047 and 62602061, Guangdong S\&T Programme (No. 2025B0101120006), the Supplemental Funds for Major Scientific Research Projects of Beijing Normal University, Zhuhai (ZHPT2023002), the Foundation of Yunnan Key Laboratory of Service Computing (YNSC26202), the Fundamental Research Funds for the Central Universities, Guangdong Provincial Education Science Planning Research Project under 2025JKZG069, and Higher Education Research Topics of Guangdong Association of Higher Education in the 14th Five-Year Plan under 24GYB207. (Corresponding author: Yanzeng Li.)}

\thanks{Haihui Liu, Yang Chen, Wenhua Wang and Yanzeng Li are with the Institute of Artificial Intelligence and Future Networks, Beijing Normal University, Zhuhai 519087, China (e-mail: haihuiliu@mail.bnu.edu.cn; yang.chen@mail.bnu.edu.cn; cs\_wwh@bnu.edu.cn; liyanzeng@bnu.edu.cn).}

\thanks{Zice Qiu is with the School of Geospatial Engineering and Science, Sun Yat-sen University, Zhuhai 519082, China (2908624009@qq.com).}

\thanks{Yun Zhang is with the Faculty of Science and Technology, Beijing Normal-Hong Kong Baptist University, Zhuhai 519087, China (v530026285@mail.bnbu.edu.cn).}

\thanks{Jialun Zhong is with the Wangxuan Institute of Computer Technology, Peking University, Beijing 100871, China (zhongjl@stu.pku.edu.cn).}

\thanks{Tian Wang is with the Engineering Research Center of Cloud-Edge Intelligent Collaboration on Big Data, Ministry of Education, Beijing Normal University, Zhuhai 519087, China (tianwang@bnu.edu.cn).}
}

\markboth{IEEE TRANSACTIONS ON KNOWLEDGE AND DATA ENGINEERING}%
{Liu \MakeLowercase{\textit{et al.}}: Graph-Mediated LLM Systems in the Agentic Era}

\maketitle

\begin{abstract}
Graphs play an increasingly important role in large language model (LLM) systems, supporting structural reasoning, knowledge grounding, retrieval, and memory. As these systems evolve toward the agentic era, graphs also organize execution dependencies and multi-agent communication, linking intermediate observations and retained information to subsequent decisions. However, understanding their contribution requires distinguishing the function of each graph operation, the component responsible for its computation, and the subsequent use of its output. This survey introduces an operation-level role--locus taxonomy for reasoning, knowledge and evidence, retrieval and memory, execution, and coordination. We compare how graph observations and external computations guide decisions, how memory representations and updates support later tasks, and how dependencies constrain execution and information delivery. These comparisons distinguish improvements in local metrics from downstream task performance while retaining the conditions under which results were reported. We then organize benchmarks, comparison designs, process diagnostics, and system costs into an evaluation framework. Finally, we identify open challenges in adaptive computation and stopping, memory maintenance, and dependency-aware recovery. The survey provides a basis for assessing when graph mechanisms are useful and for evaluating their effects across components and stages.
\end{abstract}


\section{Introduction}
\label{sec:introduction}

Large language model (LLM) systems increasingly combine generation with retrieval, external tools, and multi-step interaction~\cite{lewis2020retrieval,schick2023toolformer,yao2022react}. Their operation involves relations among facts, intermediate results, and actions, which graphs can represent explicitly. Graphs serve as reasoning objects for studying structural computation~\cite{wang2023can,tang2025grapharena}, organize knowledge and evidence for answering~\cite{pan2024unifying,peng2025graph}, and express dependencies among reasoning operations~\cite{besta2024graph}.

In agentic systems, graph operations influence decisions across steps and interactions. Retrieved evidence guides further search or tool selection~\cite{jiang2025kgagent}. Retained memory informs later tasks~\cite{xu2025amem,zhang2025gmemory}, execution dependencies govern tool calls~\cite{kim2024llmcompiler}, and communication routes determine which information reaches other agents~\cite{shen2025topologypropagation}. The analysis therefore extends from individual answers to the effects of observations, updates, and dependencies on subsequent behavior. Adaptive decision-making can use fixed graphs as well as graphs revised during interaction.

The central question is how a graph operation affects the work that follows. A local view leaves the target computation to the model~\cite{wang2026muse}, whereas delegation assigns that computation to software~\cite{wei2025graphchain}. Reusing these outputs introduces further dependencies. Consider the question, ``Which organizations operate facilities in the city where researcher A was born?'' A retrieved path provides relational evidence, and an executable query combines the relevant facts. If a facility changes operator, questions about its current and past operators require time-specific answers. A notification intended for the current operator should be addressed accordingly, even if a saved plan still names the previous one. Correct retrieval alone therefore does not ensure correct action.

\begin{figure*}[t]
\centering
\includegraphics[width=\textwidth]{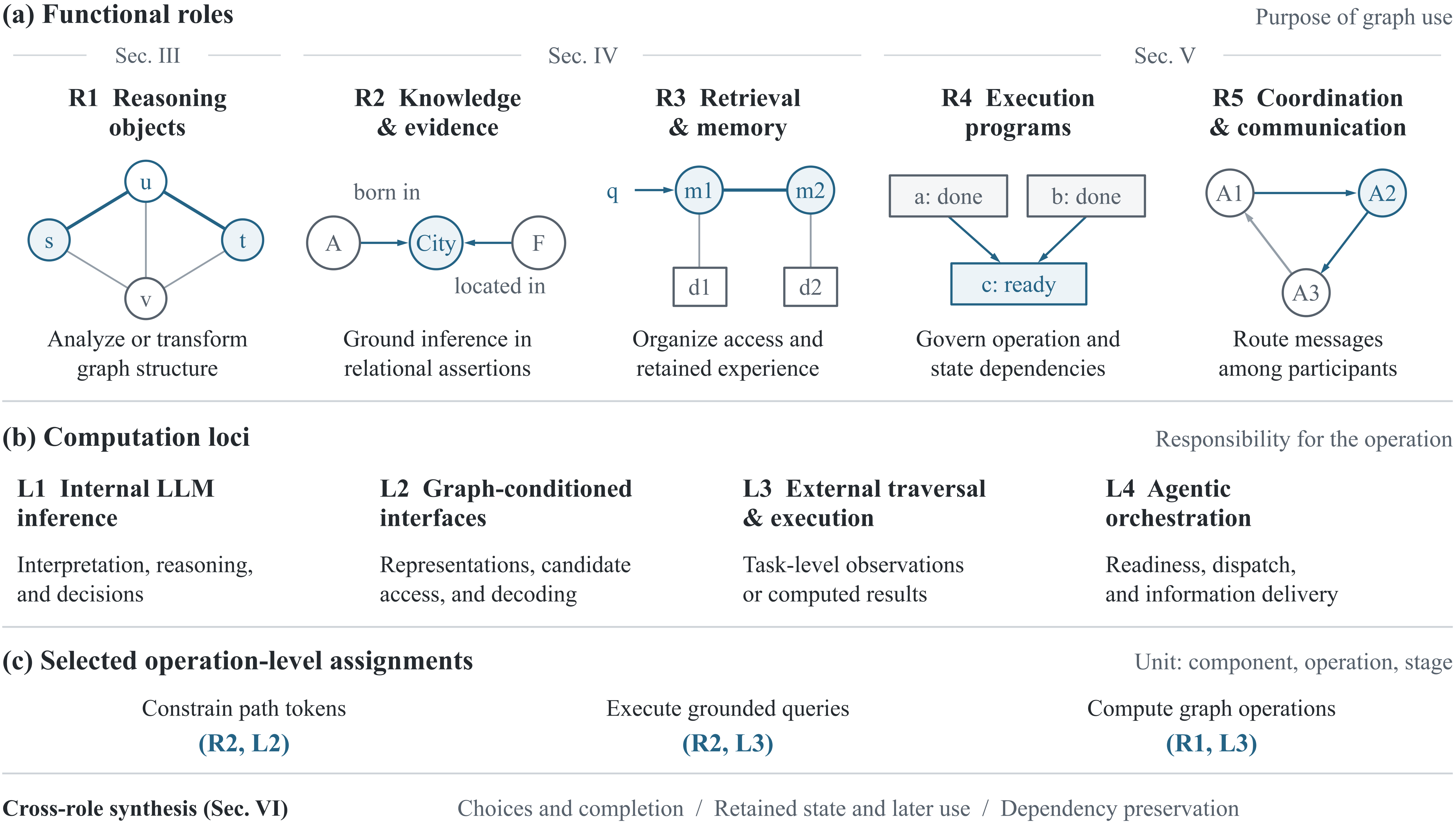}
\caption{Overview of graph roles and computation loci. (a) Role-specific examples: $s$ and $t$ are source and target vertices, with intermediate vertices $u,v$ and highlighted path $s$--$u$--$t$; $A$ and $F$ denote a person and a facility linked to a city by labeled relations; $q$ is a query, $m_1,m_2$ are memory items, and $d_1,d_2$ are associated source records; completed operations $a,b$ are prerequisites of ready operation $c$; and $A_1$--$A_3$ are agents connected by directed message routes. Blue emphasizes the illustrated path, evidence, memory links, ready operation, or message route. (b) $L_1$--$L_4$ distinguish computational responsibilities, from model inference to orchestration. (c) Selected role--locus assignments apply to individual operations; a system may contain several assignments. Section markers connect the roles to the review, and the bottom row links them to the three cross-role research questions.}
\label{fig:conceptual_overview}
\end{figure*}

Existing surveys organize graph--LLM integration by model--graph relationships~\cite{pan2024unifying}, technical roles~\cite{jin2024large}, and research directions~\cite{li2024graphmeets}. Shang and Huang~\cite{shang2025generativegraphsurvey} review graph queries, inference and learning, and applications. Other reviews cover knowledge graph (KG) question answering~\cite{ma2025large}, graph retrieval~\cite{peng2025graph}, graph-supported agents~\cite{bei2025graphsmeet,liu2026graphaugmented,luo2026graphassisted}, and workflow optimization~\cite{yue2026runtimegraphs}. Yang et al.~\cite{yang2026graphmemorysurvey} organize graph memory by its lifecycle, while Xu et al.~\cite{xu2026dynamicgraphsurvey} examine agent evolution through graph transformations. Chen et al.~\cite{chen2026agenticgraphrag} cross planner, actor, critic, and memory roles with indexing, traversal, and construction in an Agentic GraphRAG taxonomy. Our framework complements this grid by classifying stage-specific operations according to graph function and computational responsibility, including graph-object tasks, executable dependencies, and communication topology. We trace how the outputs of these operations affect subsequent decisions and how the reported experiments support those effects.

We use the term \emph{graph-mediated LLM systems} to denote graph-reasoning methods and agentic applications with identifiable operations that use graph relations. Our role--locus taxonomy separates graph function from computational responsibility. It distinguishes five roles---\emph{Reasoning Objects, Knowledge \& Evidence, Retrieval \& Memory, Execution Programs, and Coordination \& Communication}---and four computation loci. Assignments apply to individual operations at specified stages, so one system can have several assignments. Three research questions (RQs) guide comparisons across these roles:
\begin{enumerate}
    \renewcommand{\labelenumi}{\textbf{RQ\arabic{enumi}:}}
    \item\label{rq:observation} How do graph observations, externally computed results, and execution feedback guide subsequent actions and completion decisions?
    \item\label{rq:memory} How do memory representations, retrieval, and updates affect the information available for later answering, planning, and collaboration?
    \item\label{rq:dependencies} Which evidence, execution, and communication dependencies need to be preserved for correct downstream behavior?
\end{enumerate}

Our contributions are an operational taxonomy, a cross-role synthesis, and an evaluation framework. The synthesis compares computation allocation, memory access and maintenance, and dependency preservation. It highlights cases in which local gains do not translate into better downstream outcomes~\cite{wu2024longmemeval,hu2026memoryneed,mavromatis2025gnnrag,shen2025topologypropagation}. The evaluation framework connects these distinctions to task outcomes, process evidence, and system costs through whole-pipeline comparisons, conditional online comparisons, and component interventions. The remaining gaps motivate research on adaptive computation, reliable memory, and dependency-aware recovery.

Figure~\ref{fig:conceptual_overview} introduces graph roles, computational responsibilities, and selected operation-level assignments. Section~\ref{sec:scope} defines the scope and taxonomy. Sections~\ref{sec:reasoning_objects}--\ref{sec:agentic_graphs} review graph reasoning, evidence and memory, and agentic control. Section~\ref{sec:cross_cutting} synthesizes their findings around the three research questions. Section~\ref{sec:evaluation} explains how to evaluate the resulting claims, and Section~\ref{sec:challenges} identifies open challenges and directions for future research.

\section{Scope and Operational Framework}
\label{sec:scope}

\subsection{Scope and Literature Selection}
\label{subsec:scope_definition}

We include LLM systems with operations that use connectivity, relational semantics, or graph-dependent constraints in reasoning, evidence preparation, retrieval, generation, execution, or communication. We call such operations \emph{graph-active}. Offline preparation is included: Recursive Abstractive Processing for Tree-Organized Retrieval (RAPTOR)~\cite{sarthi2024raptor} constructs hierarchical summaries, and GraphRAG~\cite{edge2024local} organizes relational evidence for retrieval-augmented generation (RAG). Their prepared content can support answering without online graph traversal. We therefore distinguish graph-based preparation from graph use during answering.

We conduct a directed qualitative synthesis of work publicly available by 20 August 2026. We draw on arXiv, official proceedings, and publisher records, and use backward citation tracing to identify additional work. Searches cover graph-object reasoning, interactive knowledge graph question answering (KGQA), retrieval and memory, tool dependencies, and communication topology. Selection prioritizes distinct mechanisms, informative empirical comparisons and counterexamples, and relevant evaluation resources. Publication records and versions are matched using digital object identifiers (DOIs), base arXiv identifiers, titles, and author lists. Primary methods and experiments are distinguished from supporting baselines, benchmarks, and prior syntheses.

This selection supports a qualitative comparison of mechanisms rather than an exhaustive inventory or a pooled performance ranking. We interpret results within each study's datasets, models, access conditions, and budgets. Cross-role implications and proposed diagnostics are our synthesis of those findings; they are distinguished from the experiments reported by the cited studies. Broader memory reviews~\cite{hu2025agentmemorysurvey} and analyses of indexing, retrieval, and reading~\cite{wu2024longmemeval,hu2026memoryneed} provide supporting context.

\subsection{Role--Locus Taxonomy}
\label{subsec:taxonomy_overview}

The unit of analysis is an operation performed by a component at a specified stage in a cited paper version. \emph{Role} identifies its graph function; \emph{locus} identifies its computational responsibility. Operations within one system are classified separately. The three research questions examine downstream effects across these functional roles.

Five roles organize the literature. $R_1$, \emph{Reasoning Objects}, covers graph analysis, construction, or transformation as the task. $R_2$, \emph{Knowledge \& Evidence}, uses assertions, schemas, and executable semantics to support interpretation or constrain inference. $R_3$, \emph{Retrieval \& Memory}, organizes access to and maintenance of evidence or experience. $R_4$, \emph{Execution Programs}, represents operation and state dependencies governing execution. $R_5$, \emph{Coordination \& Communication}, organizes participant and message dependencies.

Role follows the operation's purpose. Selecting a KG view serves $R_3$; interpreting its assertions serves $R_2$. Analyzing a communication topology as a graph object serves $R_1$, while routing messages through it serves $R_5$. Plan generation and dependency-based execution are recorded at their respective stages.

Four loci describe computational responsibility. $L_1$, \emph{Internal LLM Inference}, covers structural interpretation, reasoning, and decisions performed by the model. $L_2$, \emph{Graph-Conditioned Interfaces}, covers graph-sensitive conditioning of model representations, candidate access, or decoding. $L_3$, \emph{External Graph Traversal and Execution}, performs traversal, querying, computation, or checks that return task-level observations or results. $L_4$, \emph{Agentic Orchestration}, manages readiness, dispatch, and information delivery across components. These labels distinguish computational responsibilities independently of physical deployment location or capability level. $L_2$ changes the representations or alternatives presented at a model interface; $L_3$ returns the output of a specified traversal, query, computation, or check. If external computation constructs a model-facing constraint, constructing and applying that constraint are recorded as separate operations.

Assignments follow the specified operation and output. Applying path-token constraints in Graph-Constrained Reasoning (GCR)~\cite{luo2025graphconstrained} is assigned to $R_2/L_2$, whereas executing grounded queries in ChatKBQA~\cite{luo2024chatkbqa} is assigned to $R_2/L_3$; both can be implemented by external software. MuSe's~\cite{wang2026muse} view extraction and GraphChain's~\cite{wei2025graphchain} target computation share $R_1/L_3$, but their outputs leave different work to the model. Component-level operations can also be nested within orchestration, as with result binding and dependency-based dispatch in LLMCompiler~\cite{kim2024llmcompiler}. Table~\ref{tab:role_locus_matrix} provides representative mappings.

Temporal scope and decision control further qualify role--locus assignments. A fixed graph can support adaptive exploration, while task-specific instantiation, runtime modification, and persistent memory updates involve different objects and time horizons. Offline optimization and index reuse do not themselves establish runtime graph modification or retained-state maintenance. At a specified interface, graph-based mechanisms may score, filter, mask, or gate alternatives; these mechanisms impose different restrictions and should be distinguished from checks of semantic correctness or authorization.

\begin{table*}[t]
\caption{Representative graph-sensitive operations organized by role and locus. Assignments refer to the listed operations; a study can appear on multiple branches.}
\label{tab:role_locus_matrix}
\centering
\begingroup

\newcommand{\TIGroup}[7]{%
  \node[role,draw=#2,fill=#3] (#1) at (1.62,#4) {#7};
  \draw[branch,draw=#2]
    (#1.east)--(3.46,#4)
    (3.46,#5)--(3.46,#6);}

\newcommand{\TIRow}[6]{%
  \path[fill=#2,rounded corners=1.5pt]
    (3.80,{#3-.30}) rectangle (8.72,{#3+.30});
  \draw[branch,draw=#1]
    (3.46,#3)--(3.80,#3)
    (8.72,#3)--(9.07,#3);
  \node[operation] at (3.83,#3) {#4};
  \node[locus] at (8.35,#3) {$L_{#5}$};
  \node[studies] at (9.12,#3) {#6};}

\begin{tikzpicture}[
  x=1cm,y=1cm,
  font=\fontsize{9}{10.5}\selectfont,
  text=tbInk,
  role/.style={
    draw,rounded corners=2pt,line width=.55pt,
    align=center,text width=2.92cm,
    inner xsep=4pt,inner ysep=6pt,font=\bfseries
  },
  operation/.style={
    anchor=west,align=left,text width=4.06cm,
    inner xsep=4pt,inner ysep=3pt
  },
  locus/.style={
    fill=tbHead,rounded corners=1.5pt,
    minimum width=.48cm,minimum height=.35cm,
    inner sep=2pt
  },
  studies/.style={
    anchor=west,align=left,text width=8.74cm,
    inner xsep=2pt,inner ysep=0pt,
    font=\fontsize{8.2}{9.2}\selectfont
  },
  branch/.style={line width=.55pt},
  heading/.style={
    font=\bfseries\fontsize{9.5}{11}\selectfont,
    anchor=west,inner sep=2pt
  }
]

\path[use as bounding box]
  (0,.49) rectangle (18.10,11.69);
\path[fill=tbHead]
  (0,11.09) rectangle (18.10,11.69);

\draw[tbInk,line width=.7pt]
  (0,11.69)--(18.10,11.69);
\draw[tbInk!45,line width=.4pt]
  (0,11.09)--(18.10,11.09);

\node[heading] at (.13,11.39) {Functional role};
\node[heading] at (3.83,11.39) {Operation and locus};
\node[heading] at (9.13,11.39) {Representative studies};

\TIGroup{r1}{tbBlue}{tbBlueFill}
  {10.000}{9.300}{10.700}
  {$R_1$ Reasoning\\objects}

\TIRow{tbBlue}{tbBlueFill}{10.700}
  {Reason over graph state}{1}
  {GraphWiz~\cite{chen2024graphwiz};\enspace
   G1~\cite{guo2025g1};\enspace
   GraphThought~\cite{huang2025graphthought};\enspace
   InstructGraph~\cite{wang2024instructgraph}}

\TIRow{tbBlue}{tbBlueFill}{10.000}
  {Extract local views}{3}
  {MuSe~\cite{wang2026muse}}

\TIRow{tbBlue}{tbBlueFill}{9.300}
  {Compute graph operations}{3}
  {GraphChain~\cite{wei2025graphchain};\enspace
   GraphTool-Instruction~\cite{wang2025graphtool};\enspace
   Li et al.~\cite{li2024can}}

\draw[draw=tbBlue!30,line width=.25pt]
  (3.80,8.960)--(18.10,8.960);

\TIGroup{r2}{tbRust}{tbPeachFill}
  {7.910}{7.210}{8.610}
  {$R_2$ Knowledge\\\& evidence}

\TIRow{tbRust}{tbPeachFill}{8.610}
  {Generate relation-path plans}{1}
  {RoG~\cite{luo2024reasoning};\enspace
   Plan-on-Graph~\cite{chen2024planongraph};\enspace
   VoG~\cite{zhao2026vog};\enspace
   KARPA~\cite{fang2025karpa}}

\TIRow{tbRust}{tbPeachFill}{7.910}
  {Constrain path tokens}{2}
  {GCR~\cite{luo2025graphconstrained};\enspace
   DoG~\cite{li2025decoding}}

\TIRow{tbRust}{tbPeachFill}{7.210}
  {Execute grounded queries}{3}
  {ChatKBQA~\cite{luo2024chatkbqa};\enspace
   KG-Agent~\cite{jiang2025kgagent};\enspace
   KBQA-o1~\cite{luo2025kbqa};\enspace
   CUCKOO~\cite{shim2026cuckoo}}

\draw[draw=tbRust!30,line width=.25pt]
  (3.80,6.870)--(18.10,6.870);

\TIGroup{r3}{tbRust}{tbPeachFill}
  {5.470}{4.420}{6.520}
  {$R_3$ Retrieval\\\& memory}

\TIRow{tbRust}{tbPeachFill}{6.520}
  {Compute clusters/communities}{3}
  {RAPTOR~\cite{sarthi2024raptor};\enspace
   GraphRAG~\cite{edge2024local}}

\TIRow{tbRust}{tbPeachFill}{5.820}
  {Select graph-linked candidates}{3}
  {HippoRAG~\cite{gutierrez2024hipporag};\enspace
   \mbox{GNN-RAG}~\cite{mavromatis2025gnnrag};\enspace
   ToolNet~\cite{liu2024toolnet};\enspace
   \mbox{G-Retriever}~\cite{he2024g}}

\TIRow{tbRust}{tbPeachFill}{5.120}
  {Generate memory links}{1}
  {A-MEM~\cite{xu2025amem};\enspace
   Mem0g~\cite{chhikara2025mem0};\enspace
   AriGraph~\cite{anokhin2025arigraph}}

\TIRow{tbRust}{tbPeachFill}{4.420}
  {Access linked memories}{3}
  {A-MEM~\cite{xu2025amem};\enspace
   SGMem~\cite{wu2025sgmem};\enspace
   \mbox{G-Memory}~\cite{zhang2025gmemory};\enspace
   AriGraph~\cite{anokhin2025arigraph};\enspace
   Mem0g~\cite{chhikara2025mem0}}

\draw[draw=tbRust!30,line width=.25pt]
  (3.80,4.080)--(18.10,4.080);

\TIGroup{r4}{tbGreen}{tbGreenFill}
  {3.030}{2.330}{3.730}
  {$R_4$ Execution\\programs}

\TIRow{tbGreen}{tbGreenFill}{3.730}
  {Generate workflow programs}{1}
  {AFlow~\cite{zhang2025aflow};\enspace
   WorkflowLLM~\cite{fan2025workflowllm};\enspace
   FLOW~\cite{niu2025flow}}

\TIRow{tbGreen}{tbGreenFill}{3.030}
  {Dispatch ready operations}{4}
  {GoT~\cite{besta2024graph};\enspace
   LLMCompiler~\cite{kim2024llmcompiler};\enspace
   ControlLLM~\cite{liu2023controlllm};\enspace
   GraSP~\cite{xia2026grasp}}

\TIRow{tbGreen}{tbGreenFill}{2.330}
  {Apply typed graph patches}{3}
  {GraSP~\cite{xia2026grasp}}

\draw[draw=tbGreen!30,line width=.25pt]
  (3.80,1.990)--(18.10,1.990);

\TIGroup{r5}{tbGreen}{tbGreenFill}
  {1.290}{0.940}{1.640}
  {$R_5$ Coordination\\\& communication}

\TIRow{tbGreen}{tbGreenFill}{1.640}
  {Route messages}{4}
  {G-Designer~\cite{zhang2025gdesigner};\enspace
   \mbox{EIB-LEARNER}~\cite{shen2025topologypropagation};\enspace
   AMAS~\cite{leong2025amas};\enspace
   \mbox{Graph-of-Agents}~\cite{yun2026graph}}

\TIRow{tbGreen}{tbGreenFill}{0.940}
  {Apply sparse routes/schedules}{4}
  {AgentPrune~\cite{zhang2025agentprune};\enspace
   AgentDropout~\cite{wang2025agentdropout};\enspace
   Sparse Debate~\cite{li2024sparsedebate}}

\draw[tbInk,line width=.7pt]
  (0,.49)--(18.10,.49);

\end{tikzpicture}
\endgroup
\par\smallskip
\begin{minipage}{\textwidth}
\footnotesize
\textit{Abbreviations:} GNN, graph neural network; RoG, Reasoning on Graphs; GCR, Graph-Constrained Reasoning; DoG, Decoding on Graph; GoT, Graph of Thoughts; A-MEM, Agentic Memory; SGMem, Sentence Graph Memory; EIB-LEARNER, Error-Insight Balanced Learner.
\textit{Operation scope:} The clustering/community row refers to external structure computation ($L_3$); generating the associated summaries is an LLM operation ($L_1$). Entries identify selected operations within each study.
\end{minipage}
\end{table*}

\section{Graph-Reasoning Foundations: Observation, Learning, and External Computation}
\label{sec:reasoning_objects}

Graph-object reasoning covers local queries, global constraints, optimization, and temporal traversal. These tasks differ in the information and computation they require. Feasibility, optimality, and intermediate-state correctness also need separate assessment. We review three complementary approaches: selecting graph observations, learning graph procedures, and delegating computation to external software.

\subsection{Graph Representations and Observation Selection}
\label{subsec:graph_interfaces_invariance}

Graph representations determine how relational structure is presented to the model. Encoding and descriptive order affect graph-reasoning performance~\cite{fatemi2024talk,xypolopoulos2024linearization,ge2025graphdo}. GraphInsight~\cite{cao2025graphinsight} reorganizes descriptions according to graph importance, while InstructGraph~\cite{wang2024instructgraph} combines structured verbalization with instruction tuning. Alongside input design, interactive access makes observation selection part of the reasoning process.

MuSe~\cite{wang2026muse} allows a vision--language model to request a neighborhood by specifying a root and hop radius or to produce a final answer. An external utility extracts and renders the requested view, while the model retains access to the original graph text and accumulated views. Supervised fine-tuning and reinforcement learning train the model to select views and answer questions. The utility supplies structural observations, leaving the target computation to the model.

\subsection{Graph Procedure Learning}
\label{subsec:internal_graph_reasoning}

Procedural prompting and training address the computation performed by the model. Pseudo-code prompting supplies algorithmic instructions for model inference~\cite{skianis2024pseudocode}. Learning-based approaches use different supervision signals: GraphWiz~\cite{chen2024graphwiz} combines reasoning-path supervision with preference alignment; GraphThought~\cite{huang2025graphthought} constructs action--state training trajectories; and G1~\cite{guo2025g1} uses task-specific rule- and certificate-based rewards. These approaches emphasize reasoning traces, intermediate transitions, or verifiable task outcomes.

Evaluation should distinguish individual state transitions from complete execution. The Multistep Algorithmic Reasoning Benchmark (MAGMA)~\cite{taylor2024magma} supplies correct preceding states during stepwise evaluation, and its reported trajectory accuracy averages correctness across steps. In some breadth-first search (BFS) and depth-first search (DFS) settings, better intermediate predictions do not yield corresponding final-step gains. These results motivate separate assessment of transitions from valid supplied states and complete runs with model-generated intermediate states.

Transfer provides another test of learned procedures. G1 reports positive transfer across its evaluated graph tasks and encodings~\cite{guo2025g1}. Zhang et al.~\cite{zhang2026generalizable} report advantages for process-based rewards on synthetic graph tasks, but neither process nor answer rewards consistently dominate on downstream tasks with implicit relational structure. Graph-learning results therefore need to be interpreted in relation to the training tasks, representations, and transfer targets.

\subsection{External Graph Computation}
\label{subsec:external_graph_execution}

Delegation assigns selected graph computations to external software. Graph-library interfaces~\cite{li2024can,wang2025graphtool} reduce model-side state tracking, while the system must still identify the correct graph, operation, and arguments. GraphChain~\cite{wei2025graphchain} chains specialized tools, stores intermediate results externally, and returns compact descriptions for subsequent decisions. Its tools compute requested paths or flows, whereas the view extractor in MuSe~\cite{wang2026muse} supplies local observations for model reasoning. GraphChain's learned policy selects and composes operations, shifting the focus from internal state computation to specifying and using external results.

The benefit of external assistance depends on task conditions. GraphArena~\cite{tang2025grapharena} distinguishes invalid, feasible, and optimal solutions. In its GPT-4o~\cite{openai2024gpt4o} comparison across polynomial-time tasks, code assistance lowers accuracy on small graphs but improves it on large graphs. The authors suggest extraction and code-generation errors as possible explanations for the accuracy loss on small graphs. This contrast highlights the importance of translating task descriptions into executable computations. GraphEval36K~\cite{wu2025grapheval36k} complements answer-level measures by running generated programs on graph instances, assessing functional correctness on the tested inputs.

These foundations also apply to graphs that organize agent behavior. WorFBench~\cite{qiao2025benchmarking} evaluates generated workflow structures, Follow the Flow~\cite{suri2025followflow} uses symbolic flowchart traversal for attribution, and HARNESSFIX~\cite{chen2026harnessfix} analyzes failed-trajectory dependencies to locate editable harness logic. These structures can be analyzed as graph objects or used to govern execution through their operation dependencies.

Observation design, procedure learning, and delegation can operate together, determining which relations the model sees and which computations it performs. Section~\ref{sec:graphrag} extends this analysis to evidence access and retained memory. Table~\ref{tab:benchmark_landscape} summarizes the evaluation resources used throughout the survey.

\section{Evidence and Memory: Access, Interpretation, and Retained State}
\label{sec:graphrag}
\label{sec:kg_reasoning}

Graphs organize access to evidence and support interpretation of its assertions. Construction and retrieval determine which information reaches the model; paths and executable queries specify how relations support an answer. Sections~\ref{subsec:graphrag_construction} and~\ref{subsec:graphrag_retrieval} examine evidence preparation and selection. Sections~\ref{subsec:kg_path_search}--\ref{subsec:kg_hybrid_open_world} cover interpretation, generation constraints, and incomplete evidence. Sections~\ref{subsec:graphrag_memory} and~\ref{subsec:memory_consumers} extend the analysis to information retained across interactions and reused for action or collaboration.

\subsection{Graph Construction and Evidence Access}
\label{subsec:graphrag_construction}

Graph construction shapes both searchable representations and answering content. Passage nodes preserve source text, entities connect mentions, and summaries compress related records. GraphRAG~\cite{edge2024local} constructs entity communities and reports for global map--reduce answering. Community reports and graph-free source summaries both improve judged comprehensiveness and diversity over vector retrieval. The additional benefit of community reports varies with community resolution and evaluation protocol; the claim-based analysis finds no significant difference between community reports and graph-free summaries on these measures. RAPTOR~\cite{sarthi2024raptor} constructs recursive summaries and retrieves across hierarchy levels as a collapsed collection in its main experiments. These examples distinguish graph-based content preparation from online structural access.

G-Retriever~\cite{he2024g} combines connected retrieval with graph-conditioned generation. Prize-collecting Steiner tree (PCST) optimization selects a subgraph, which provides textual context and is encoded by a graph neural network (GNN) to produce a soft prompt. In its WebQuestionsSP (WebQSP)~\cite{yih2016webqsp} ablation, replacing the graph encoder with unstructured soft tokens or removing textualized context lowers Hit@1. This supports complementary input channels. Here, the graph-conditioned soft prompt serves as a model interface, whereas PCST-based selection performs external subgraph computation. The separate selector comparisons also change retrieved content, so they do not isolate connectivity.

Other approaches combine graph-based evidence selection with textual or graph representations. Graph Retrieval-Augmented Generation (GRAG)~\cite{hu2025grag} combines pruned ego-graphs with hierarchical text and graph prompts, while SubgraphRAG~\cite{li2025subgraphrag} scores triples using directional-distance features. LightRAG~\cite{guo2025lightrag} links entity and relation descriptions to neighboring information and source chunks. Its reported incremental insertion mechanism extends the index; replacement and deletion require separately defined update semantics.

\subsection{Retrieval Objectives and Evidence Connectivity}
\label{subsec:graphrag_retrieval}

Relational retrieval connects facts that may be weak query matches individually. HippoRAG~\cite{gutierrez2024hipporag} maps query entities to phrase nodes and propagates personalized PageRank (PPR) scores to rank passages. Its ablations favor PPR over seed-only or immediate-neighbor access. At a top-5 cutoff, its all-support recall exceeds that of ColBERTv2~\cite{santhanam2022colbertv2} on MuSiQue~\cite{trivedi2022musique} and 2WikiMultiHopQA (2Wiki)~\cite{ho2020twowiki}, but not HotpotQA~\cite{yang2018hotpotqa}. Query-Aware Flow Diffusion RAG (QAFD-RAG)~\cite{zhou2026qafd} uses query-weighted, capacity-constrained flow, while Structure-Tracing Evidence Mining (STEM)~\cite{yu2026stem} guides subgraph search with a question-derived schema and triple-dependent signals.

Selection quality depends on how retrieved relations are used. GNN-RAG~\cite{mavromatis2025gnnrag} predicts answer candidates with a question-conditioned GNN and extracts shortest paths for LLM answering. In experiments using pre-extracted candidate subgraphs, sparsification improves GNN node prediction on ComplexWebQuestions (CWQ)~\cite{talmor2018complexwebquestions} but lowers downstream LLM answering performance. The authors identify disconnected support that prevents path extraction. This contrast highlights the importance of preserving paths between question entities and candidate answers.

Learning can also target the usefulness of retrieved evidence. Knowledge Graph-Enhanced Large Language Models via Path Selection (KELP)~\cite{liu2024kelp} supervises paths through answering utility; Retrieval-and-Reasoning~\cite{ji2024retrievalreasoning} trains retrieval and KG-aware reasoning separately. RAPL~\cite{yao2025rapl} combines a rationalized annotator, path-based reasoning, and a line graph of triples. With linked question entities and answer-containing candidates, replacing its retrieved paths with shuffled triples lowers CWQ Macro-F1 for two GPT readers. This result supports the path representation as a whole. Because grouping and order change together, the comparison does not isolate either factor.

Generalization depends on the evaluated shift. Graph Foundation Model for Retrieval-Augmented Generation (GFM-RAG)~\cite{luo2025gfmrag} maps query-conditioned entity scores to documents and reports transfer without domain-specific fine-tuning. G-Reasoner~\cite{luo2026greasoner} retrieves over document, KG, attribute, and community nodes, with graph-constructor families represented in training. On MuSiQue-Full, G-Reasoner achieves higher answer exact-match and F1 scores than the Qwen3-Emb-8B dense-retrieval baseline, whereas HippoRAG~\cite{gutierrez2024hipporag} and GFM-RAG~\cite{luo2025gfmrag} score lower. These evaluations probe different aspects of generalization: domain shift, exposure to graph constructors, and corpus expansion.

\subsection{KGQA: Interpretation, Feedback, and Completion}
\label{subsec:kg_path_search}
\label{subsec:kg_semantic_parsing}
\label{subsec:unified_graph_control_loop}

KGQA requires an interpretation that specifies how relations answer the question. Reasoning on Graphs (RoG)~\cite{luo2024reasoning} predicts relation plans and retrieves their KG instances. ChatKBQA~\cite{luo2024chatkbqa} generates a compositional S-expression, grounds identifiers, and executes queries in SPARQL (SPARQL Protocol and RDF Query Language)~\cite{harris2013sparql} for joins, counting, and extrema. Here, RDF denotes the Resource Description Framework. In ChatKBQA's non-oracle WebQSP analysis, logical-skeleton errors form the largest error category, highlighting the importance of interpretation before execution. SELF-KBQA~\cite{tian2026selfkbqa}, GRV-KBQA~\cite{tian2025grvkbqa}, and DR2~\cite{liang2026diagnosing} respectively target logic-compatible evidence selection, structural query validation, and relation representations. SEAL~\cite{wang2026seal} addresses conversational KGQA with calibrated S-expression cores, template-based completion, and local/global memories for feedback-driven adaptation.

Interactive methods use partial evidence to guide further search. StructGPT~\cite{jiang2023structgpt} alternates structured-data access and model reasoning. Think-on-Graph~\cite{sun2024think} explores relations and entities on a fixed KG, choosing whether to continue or answer; its depth-limit fallback can use parametric knowledge. Plan-on-Graph~\cite{chen2024planongraph} tracks paths and sub-objectives, using reflection to expand or backtrack. Removing reflection lowers Hits@1 on CWQ, WebQSP, and sampled GrailQA~\cite{gu2021grailqa} under its annotated-topic-entity protocol. Verify-on-Graph (VoG)~\cite{zhao2026vog} instead revises plans when predicted observations conflict with retrieved facts.

Feedback can also revise executable interpretations. KBQA-o1~\cite{luo2025kbqa} searches executable logical-form operations, while DAMR~\cite{wang2026damr} evaluates partial relation paths. CUCKOO~\cite{shim2026cuckoo} refines grounded logical forms after empty results; Execution as Verification~\cite{zhang2026executionverification} combines atomic execution checks with semantic alignment and rollback. An empty result may be the correct answer to the intended question, or it may reflect an unsuitable interpretation or missing knowledge. Execution feedback therefore needs semantic assessment.

Task-local memory retains information for subsequent choices. In KG-Agent~\cite{jiang2025kgagent}, the model selects functions and arguments, an executor performs KG operations, and task memory stores retrieved information and program history. Graph Chain-of-Thought (Graph-CoT)~\cite{jin2024graphcot} accesses text on attributed graphs; RefKG~\cite{zhou2025reflection} revises retrieved chains and reconstructs evidence. ARK~\cite{polonuer2026autonomous} records retrieved nodes and returns a ranked list when it selects \texttt{FINISH} or reaches a step limit. Retrieval completion and answer completion thus concern different outputs.

Graph Explorer~\cite{chen2026graphexplorer} connects action visibility with graph-supported completion. Offline tool-plan supervision is derived from expert SPARQL queries and filtered according to argument visibility. At inference, the model selects actions over a fixed KG, with intermediate sets retained in a registry. For its Qwen3-8B~\cite{yang2025qwen3} model trained only on CWQ, masking expert-plan hop relations from visible candidates sharply reduces finish-or-fail Hit@1 on the reported CWQ subset, while the backend KG remains unchanged. The contrast concerns access to actionable relations; runtime reference and execution checks remain distinct from its post-hoc visibility audit.

Search allocation also affects interaction cost. KARPA~\cite{fang2025karpa} combines planned relations with matching and revision, while Chain-of-Relations (CoR)~\cite{liu2026chainrelations} delays global entity filtering until relation-chain construction. Both report fewer interactions than their selected baselines. Section~\ref{subsec:workflow_graphs} examines feedback that revises workflow programs, complementing the task-local evidence and interpretation updates considered here.

\subsection{Constrained Generation and Verification Boundaries}
\label{subsec:kg_constrained_verification}

Constrained generation makes selected structural properties checkable. GCR~\cite{luo2025graphconstrained} builds a KG-Trie from bounded-hop paths and masks invalid token continuations. Completed constrained paths belong to the indexed language, while hypothesis generation and answer aggregation are not subject to that path-token constraint. On CWQ, removing the constraint sharply lowers path conformance among correctly answered questions while answer accuracy changes little. The conditional conformance measure and answer accuracy therefore assess different properties.

Decoding on Graph (DoG)~\cite{li2025decoding} updates a query-centric graph view and trie during generation. FiDeLiS~\cite{sui2025fidelis} combines relevance and connectivity in constrained candidates, while OntGQA~\cite{shan2026ontgqa} uses ontology-compatible types and relations. Conformal Path Reasoning~\cite{lin2026conformalpath} calibrates path sets under exchangeability assumptions. Path membership, type compatibility, and statistical coverage provide different checks; alignment with the intended question remains a separate requirement.

\subsection{Incomplete and Mixed Evidence}
\label{subsec:kg_hybrid_open_world}

Incomplete KGs motivate external evidence retrieval and knowledge completion. Under open-world assumptions, an empty query result does not establish real-world falsity. Think-on-Graph~2.0~\cite{ma2025thinkongraph2} consults topic-entity text before graph expansion and uses later text to prune candidates. Its restricted-discovery experiment on sampled HotpotQA questions reports lower performance when fewer entities and relations are exposed, with partial recovery after search adjustment. Generate-on-Graph~\cite{xu2024generateongraph} and joint knowledge-base completion and question answering~\cite{liu2026jointkbcqa} incorporate proposed knowledge into answering.

GLOW~\cite{abdallah2026glow} combines predicted answer candidates with SPARQL-retrieved neighborhood facts. Its protocol excludes the gold answer from returned context, and some two-hop settings favor facts alone over facts plus predictions. This context restriction does not establish that supporting paths are absent from the KG. MindMap~\cite{wen2024mindmap} and Right for Right Reasons~\cite{toroghi2024right} also combine graph evidence with model knowledge or surfaced commonsense assumptions. GAKI~\cite{zhong2027gaki} constructs a patient-centric graph from dialogue and combines it with external medical knowledge to support medication recommendation. These sources should remain distinguishable when conclusions become stored facts or action arguments.

\subsection{Dialogue Memory: Access, and Maintenance}
\label{subsec:graphrag_memory}

A central goal of dialogue memory is to retain useful information and incorporate changes across interactions. HippoRAG~2~\cite{gutierrez2025hipporag2} evaluates associative retrieval under corpus expansion; persistent dialogue use additionally requires handling revisions to retained information. The key choices concern search representations, returned content, and maintenance semantics.

LongMemEval~\cite{wu2024longmemeval} separates indexing, retrieval, and reading. In its budget-aware LongMemEval$_M$ analysis, using extracted facts as returned values lowers overall question answering (QA) performance but improves multi-session reasoning. The benefit of decomposing sessions into rounds also depends on the reader. These results motivate separate decisions about retained sources, retrieval units, and the context supplied for answering.

Hu et al.~\cite{hu2026memoryneed} further distinguish keys, values, queries, indexing, retrieval, and answering. Their Table~7 reports that selected graph configurations outperform flat memory with original-session values but underperform it when derived keys are returned. The evaluated subsets exclude questions whose reference answers occur outside user turns; session- and key-based retrieval also provide different amounts of source detail. The result depends on the returned-value configuration, leaving the contribution of topology unresolved.\footnote{Table~7 labels the answering model accessed through the application programming interface (API) as GPT-4o, whereas Appendices~C.1 and D.2 use GPT-4o-mini~\cite{openai2024gpt4omini}. This discrepancy precludes API-model-specific conclusions from the comparison.} The reported no-expansion configuration still uses entity alignment and source mapping to rank returned values. Graph-based preparation can thus support evidence recovery without neighbor traversal at query time.

SGMem~\cite{wu2025sgmem} recovers context through sentence-similarity expansion and membership links to parent chunks. Its construction avoids LLM-based entity extraction, although summaries and facts still require generation. Comparisons with chunk-based RAG variants and dataset-dependent hop sensitivity support selective source recovery, complementing the separation between searchable representations and answering context.

Maintenance has effects beyond extraction quality. In Hu et al.'s~\cite{hu2026memoryneed} flat-memory HaluMem-Medium~\cite{chen2025halumem} comparison, jointly enabling Add/Update/Noop improves recall and QA over Add alone but lowers target precision. The comparison does not isolate Update from Noop and does not test deletion. Graph configurations separately combine higher extraction precision with lower recall and backbone-dependent QA. These findings distinguish extraction quality from the support available for later answers.

Systems implement different update semantics. A-MEM~\cite{xu2025amem} links contextualized notes and evolves earlier notes during ingestion. Mem0~\cite{chhikara2025mem0} supports addition, modification, deletion, and no change; Mem0$^g$ invalidates conflicting edges while retaining history. On LoCoMo~\cite{maharana2024locomo}, the graph variant improves temporal and open-domain judge scores but trails flat memory on multi-hop questions, with higher search and response latency. Representation, retrieval, and maintenance change together in this comparison. Zep~\cite{rasmussen2025zep} distinguishes fact-validity time from ingestion and invalidation time, supporting temporally qualified access to retained information.

Non-graph systems also support memory maintenance. Reflective Memory Management (RMM)~\cite{tan2025rmm} combines topic summaries with feedback-informed reranking. LightMem~\cite{fang2025lightmem} appends entries online and consolidates them offline. It reports QA performance both before and after offline updates on LongMemEval$_S$, whereas its LoCoMo accuracy is measured after the update. Maintenance timing determines which state later questions can use. A formerly valid assertion may remain relevant to a historical question, while an explicit deletion request concerns retention itself.

\subsection{Memory for Action and Collaboration}
\label{subsec:memory_consumers}

Action-oriented memory supplies state and experience for environmental decisions. AriGraph~\cite{anokhin2025arigraph} updates semantic triples from observations and links them to episodic records. Retrieved memory informs planning and action. In TextWorld~\cite{cote2018textworld}, episodic recall helps recover cooking instructions, whereas cleaning tasks emphasize correcting stale object locations. These results concern state use within the evaluated runs; Table~\ref{tab:graphrag_methods} records the aggregation protocol.

G-Memory~\cite{zhang2025gmemory} retains past queries, collaboration traces, and abstract insights. Retrieval provides role-specific guidance at task onset, and the store is updated after task completion. Its retrieval-hop comparisons across tasks and teams show that wider expansion can lower performance. The experience graph supports retrieval of guidance from earlier tasks; the active communication topology determines message delivery during the current task.

Procedural experience supplies another form of guidance. Thought-Action Graph Reasoning~\cite{qi2026thought} retrieves abstract operation chains and corresponding KGQA actions, while Reflexion~\cite{shinn2023reflexion} reuses verbal evaluations without explicit graph memory. Table~\ref{tab:graphrag_methods} compares representative memory mechanisms and their tested uses. Section~\ref{sec:cross_cutting} synthesizes how factual evidence, applicable state, and procedural experience support subsequent decisions.

\begin{table*}[t]
\caption{Memory mechanisms compared by content, maintenance, and subsequent use. Empirical contrasts retain their study-specific conditions.}
\label{tab:graphrag_methods}
\centering
\begingroup
\small
\setlength{\tabcolsep}{5pt}
\setlength{\fboxsep}{3pt}
\renewcommand{\arraystretch}{1.10}
\begin{tabularx}{\textwidth}{
  @{}
  >{\raggedright\arraybackslash}p{0.135\textwidth}
  >{\raggedright\arraybackslash}p{0.25\textwidth}
  >{\raggedright\arraybackslash}p{0.27\textwidth}
  >{\raggedright\arraybackslash}X
  @{}
}
\toprule
\TableHeader{Study}
& \TableHeader{Content and access}
& \TableHeader{Maintenance and later use}
& \TableHeader{Evidence and conditions} \\
\midrule
\multicolumn{4}{@{}l@{}}{\TableBand{tbPeachFill}{tbRust}{Dialogue memory}} \\
\addlinespace[3pt]
\textbf{Hu et al.}~\cite{hu2026memoryneed}
& Derived keys or entity descriptions; source alignment and value ranking.
& Session ingestion; original sessions or derived keys supplied for answering.
& Graph/flat ordering reverses across returned-value configurations on filtered question subsets. \\
\addlinespace[4pt]
\textbf{Mem0 / Mem0$^g$}~\cite{chhikara2025mem0}
& Facts or entity relations; semantic and relational retrieval.
& Mem0 modifies or deletes facts; Mem0$^g$ invalidates edges for later dialogue QA.
& LoCoMo category gains and losses; graph memory has higher search and response latency. \\
\addlinespace[4pt]
\textbf{A-MEM}~\cite{xu2025amem}
& Contextualized notes and association links.
& Earlier notes are revised during ingestion to support subsequent retrieval.
& The full system outperforms variants without evolution or without both linking and evolution; no ablation removes linking alone. \\
\addlinespace[4pt]
\textbf{LightMem}~\cite{fang2025lightmem}
& Compressed, topic-organized entries; semantic retrieval.
& Online append and offline consolidation for later QA.
& LongMemEval$_S$: pre- and post-update QA; LoCoMo: post-update QA. Construction costs include summarization and updating. \\
\addlinespace[5pt]
\multicolumn{4}{@{}l@{}}{\TableBand{tbPeachFill}{tbRust}{Action and team memory}} \\
\addlinespace[3pt]
\textbf{AriGraph}~\cite{anokhin2025arigraph}
& Semantic triples linked to full-observation episodes.
& Observation-time revision supports planning and action within the environment.
& TextWorld cooking/cleaning comparisons average the best three of five attempts per agent/game. \\
\addlinespace[4pt]
\textbf{G-Memory}~\cite{zhang2025gmemory}
& Query, interaction, and insight graphs; role-specific experience retrieval.
& Task-end assimilation supplies guidance to subsequent teams.
& Adapted action/QA protocols; insight/trace ablations on Planning Domain Definition Language (PDDL) tasks from AgentBoard~\cite{ma2024agentboard} and Fact Extraction and VERification (FEVER)~\cite{thorne2018fever}; task/team hop sensitivity. \\
\bottomrule
\end{tabularx}
\endgroup
\end{table*}

\section{Agentic Control: Executable Dependencies and Communication}
\label{sec:agentic_graphs}

Agentic control determines which operations proceed and which information reaches each decision-maker. Execution graphs encode operation and state dependencies; communication graphs organize participation and message delivery. We first examine workflow representation and dispatch, then tool composition and execution checks, and finally communication topology. This progression distinguishes planned dependencies, runtime enforcement, and information flow.

\subsection{Plan Representations and Dependency-Guided Execution}
\label{subsec:workflow_graphs}

Graph-based plans can support reasoning about execution or directly govern it. Plan Like a Graph (PLaG)~\cite{lin2024plag} represents ordering constraints and evaluates minimum completion time under unlimited resources; actual dispatch lies outside its AsyncHow evaluation. MatrixCoT~\cite{chen2026matrixplan} represents logical-step dependencies with an explicit matrix and revises the reasoning plan using execution-trace feedback, without relying on external solvers. Graph of Thoughts (GoT)~\cite{besta2024graph} executes generation, scoring, selection, and aggregation operations according to their dependencies. Its Graph of Operations (GoO) specifies the structure, while the Graph Reasoning State (GRS) records thoughts and assessments. The state can evolve under a fixed operation graph.

LLMCompiler~\cite{kim2024llmcompiler} turns call dependencies into asynchronous execution by binding arguments to completed outputs and releasing ready calls. In Game of~24~\cite{yao2023tree}, evaluated states return to the planner; selected states seed another plan when no proposal reaches 24. Parallel candidate generation and evaluation reduce latency relative to sequential Tree of Thoughts~\cite{yao2023tree} in that comparison. On WebShop~\cite{yao2022webshop}, however, LLMCompiler achieves a higher success rate than ReAct~\cite{yao2022react} with both evaluated GPT variants, but also incurs greater latency, which the authors attribute to planner overhead. These results illustrate workload-dependent benefits and costs of the complete workflow. At serving scale, HeraSys~\cite{li2026herasys} merges reusable nodes across concurrent workflows and schedules work both within and across requests, targeting throughput and tail latency.

Workflow construction also differs in how programs are obtained. GPTSwarm~\cite{zhuge2024gptswarm} optimizes component connections while preserving required internal edges and acyclicity; node optimization changes prompts. AFlow~\cite{zhang2025aflow} uses execution results on validation tasks to search over programs containing operators, branches, and loops before deployment. Its reported deployment-execution costs exclude the preceding search. WorkflowLLM~\cite{fan2025workflowllm} instead learns hierarchical workflows from converted shortcuts and synthesized examples.

Execution representations determine what can be revised. Production workflow systems~\cite{park2025workflowgraphs} expose specified routes, history transformations, and typed interfaces; Agent Spec~\cite{amini2026agentspec} separates control-flow and data-flow edges for cross-framework execution. FLOW~\cite{niu2025flow} supports runtime refinement of an activity-on-vertex graph using subtask progress and execution feedback. These mechanisms distinguish program construction from changes to deployed dependencies.

\subsection{Tool Selection and Resource-Compatible Composition}
\label{subsec:tool_access_composition}

Tool composition involves selecting candidates, matching resources, and binding calls. ToolNet~\cite{liu2024toolnet} retrieves successors of the previous tool and presents transition weights as selection preferences. The model chooses a tool and its arguments, then observes the result. ToolNet updates transition preferences from completed trajectories; its simulated API-failure experiment illustrates adaptation toward functioning alternatives. These links guide tool access without certifying invocation conditions.

ControlLLM's Thoughts-on-Graph~\cite{liu2023controlllm} searches a bipartite graph linking tools to accepted and produced resource types. The search tracks available resources, while LLM components select plans and resolve bindings. The execution engine follows the selected plan and can parallelize independent subtasks. GTool~\cite{chen2026gtool} constructs request-specific graphs from historical relations and predicts missing dependencies for planning. Resource-compatible paths and predicted links support composition; reliable execution also calls for checking current inputs and state conditions.

\subsection{Execution Checks and Runtime Repair}
\label{subsec:tool_skill_graphs}

Skill retrieval supplies candidate procedures. Skill Retrieval Augmentation~\cite{su2026skillretrieval} distinguishes making a skill available from actually using it, while SkillComposer~\cite{zhao2026skillcomposition} selects and loads relevant skill sequences, using dependencies during offline synthesis. Checked composition additionally evaluates whether invocations are applicable and whether their expected effects occur.

GraSP~\cite{xia2026grasp} compiles skills into a typed directed acyclic graph (DAG) of invocations, bound arguments, and verifiers. State and data dependencies constrain execution, while order edges can express softer precedence. Preconditions are checked before invocation and postconditions afterward. Failures trigger repairs such as rebinding, prerequisite insertion, or rewiring. Patches are bounded by size and graph neighborhood, with escalation to global replanning or reactive fallback when local repair fails. 

The skill-augmented systems compared in GraSP use the same skill libraries and episodic memory. With DeepSeek-V3.2~\cite{deepseek2025v32} on seen tasks in ALFWorld~\cite{shridhar2021alfworld} and ScienceWorld~\cite{wang2022scienceworld}, the sequential-execution and global-replanning variants score below the full system on the success and reward measures for those environments~\cite{xia2026grasp}. The local-versus-global repair comparison matches repair-attempt budgets, which does not establish equal token or environment-action costs. Postcondition checks can return uncertain LLM judgments, so their verdicts remain conditional on represented state and available observations. In a domain-specific multi-agent system, RareDASH~\cite{zhong2026raredash} uses an auditing agent to trigger node re-execution or DAG restructuring; its demo does not isolate the effect of repair.

Runtime recovery also depends on observed environment state. Agent-SAMA~\cite{guo2026agentsama} constructs mobile-application state graphs, compares expected transitions with observed screens, and uses state information for recovery. Retained experience can inform later tasks. Feedback is also available under structured protocols: MetaGPT~\cite{hong2024metagpt} uses prescribed procedures and execution results, while AutoGen~\cite{wu2024autogen} supports programmable conversations among models, tools, and human participants. Such systems provide comparisons for assessing the additional value of structural adaptation.

\subsection{Communication Topology and Information Flow}
\label{subsec:communication_topologies}

Communication graphs specify who participates and which messages reach other components. Removing routes can reduce redundant context while blocking information needed for a decision. Self-consistency~\cite{wang2022self}, which aggregates independent reasoning samples, provides a comparison without inter-agent exchange. GPTSwarm~\cite{zhuge2024gptswarm} reports recovery toward single-agent performance through edge optimization in an adversarial Massive Multitask Language Understanding (MMLU)~\cite{hendrycks2021mmlu} setting, and gains in a separate collaborative setting with diverse role prompts. The adversarial and collaborative results use different participant settings and should be reported separately when assessing topology effects.

Topology selection can precede interaction. G-Designer~\cite{zhang2025gdesigner} generates edges using agent and tool profiles with task information; Adaptive Multi-Agent System (AMAS)~\cite{leong2025amas} selects among optimized candidate graphs. Guided Topology Diffusion (GTD)~\cite{jiang2026gtd} uses graph diffusion with a learned utility--cost proxy, while TopoDIM~\cite{sun2026topodim} samples typed interactions under structural constraints. These mechanisms condition route selection on information available before the conversation.

Pruning differs in both timing and the structure selected. AgentPrune~\cite{zhang2025agentprune} learns masks and deletes edges once, either after early rounds of one query or after optimization on a query subset for later reuse. AgentDropout~\cite{wang2025agentdropout} learns round-specific participation and samples a complete multi-round schedule before reasoning. Reported trade-offs include lower prompt-token use with a small MMLU accuracy loss for AgentPrune on GPTSwarm, and lower average scores at higher AgentDropout rates. Deployment reports should include optimization costs and specify how many queries reuse the resulting schedule or topology.

Routing interacts with aggregation and shared state. Sparse-debate experiments~\cite{li2024sparsedebate} compare sparse neighbor communication with dense exchange; Graph-of-Agents~\cite{yun2026graph} combines agent and edge sampling with message passing and pooling. AgentNet~\cite{yang2025agentnet} separates routing and execution using experience buffers, while EvoHyper~\cite{zhang2026evohyper} connects groups through shared memory and distinguishes memory updates from structural changes.

EIB-LEARNER~\cite{shen2025topologypropagation} examines how topology affects the propagation of supplied information. Its MMLU protocol induces local errors on questions initially answered correctly and injects gold answers on questions initially answered incorrectly. These correctness-conditioned question sets are selected separately for each topology. Degree-normalized answer changes indicate that sparse routes reduce the influence of both harmful and useful supplied information, whereas denser routes increase their influence. These interventions measure how supplied information changes answers; the discovery of corrections falls outside their scope. The learned method combines sparse and dense views through query-aware fusion and selects a topology before interaction; branch and fusion ablations support this design under a separate task-evaluation protocol.

Communication design also needs to account for message attacks and propagation failures. NetSafe~\cite{yu2025netsafe} reports task-dependent resilience under its one-attacker protocol: chains perform better on factual and commonsense subsets, while complete communication is favored on mathematics. Agent-in-the-Middle (AiTM)~\cite{he2025aitm} attacks inter-agent messages, Communication Inference Attack (CIA)~\cite{wu2026cia} infers communication links from outputs, and GUARDIAN~\cite{zhou2025guardian} uses temporal interaction graphs to detect propagated anomalies. These studies examine message attacks, topology exposure, and anomaly detection alongside task performance.

Table~\ref{tab:agentic_control_methods} connects control mechanisms to their decision inputs and timing. Execution comparisons concern readiness and recovery; communication comparisons concern whether information reaches and influences a decision-maker. Section~\ref{sec:cross_cutting} synthesizes these requirements with evidence access and retained memory.

\begin{table*}[t]
\caption{Feedback and selection in graph-based control. Each row identifies the relation used, the decision input, and the timing of selection or revision.}
\label{tab:agentic_control_methods}
\centering
\begingroup
\small
\setlength{\tabcolsep}{5pt}
\setlength{\fboxsep}{3pt}
\renewcommand{\arraystretch}{1.10}

\begin{tabularx}{\textwidth}{
  @{}
  >{\raggedright\arraybackslash}p{0.205\textwidth}
  >{\raggedright\arraybackslash}p{0.22\textwidth}
  >{\raggedright\arraybackslash}p{0.235\textwidth}
  >{\raggedright\arraybackslash}X
  @{}
}
\toprule
\TableHeader{Method and relation}
& \TableHeader{Feedback or selection input}
& \TableHeader{Affected decision}
& \TableHeader{Timing and scope} \\
\midrule

\multicolumn{4}{@{}l@{}}{
  \TableBand{tbGreenFill}{tbGreen}{Computation and execution}
} \\
\addlinespace[3pt]

\textbf{GoT}~\cite{besta2024graph}
\newline Operation dependencies
& Thought scores and validity checks.
& Select, aggregate, or refine thoughts.
& Task-local GRS updates under a specified GoO. \\
\addlinespace[4pt]

\textbf{LLMCompiler}~\cite{kim2024llmcompiler}
\newline Call dependencies
& Completed outputs; evaluated search states.
& Bind arguments, release ready calls, and replan.
& Runtime dispatch; reaching 24 completes the Game of 24 search. \\
\addlinespace[4pt]

\textbf{AFlow}~\cite{zhang2025aflow}
\newline Workflow program
& Validation execution and scores.
& Revise prompts, operators, and dependencies.
& Search before deployment; the selected program may branch or loop. \\
\addlinespace[4pt]

\textbf{GraSP}~\cite{xia2026grasp}
\newline Typed skill dependencies
& Failed precondition, execution, or postcondition checks.
& Apply bounded patches; escalate when needed.
& Runtime repair; checker verdicts may be uncertain. \\
\addlinespace[5pt]

\multicolumn{4}{@{}l@{}}{
  \TableBand{tbGreenFill}{tbGreen}{Communication selection}
} \\
\addlinespace[3pt]

\textbf{AgentPrune}~\cite{zhang2025agentprune}
\newline Message edges
& Task utility during mask learning.
& Learn masks, then prune edges once.
& Early rounds of one query or a training query subset; subsequent topology fixed. \\
\addlinespace[4pt]

\textbf{AgentDropout}~\cite{wang2025agentdropout}
\newline Participation schedule
& Learned participation parameters.
& Sample a complete multi-round schedule.
& Before reasoning; participation varies across rounds. \\
\addlinespace[4pt]

\textbf{EIB-LEARNER}~\cite{shen2025topologypropagation}
\newline Message routes
& Task query and learned sparse/dense fusion.
& Select a communication topology.
& Before interaction; supplied-information propagation is evaluated separately. \\

\bottomrule
\end{tabularx}
\endgroup
\end{table*}

\begin{table*}[t]
\caption{Cross-role answers and their evidence basis. Design implications and diagnostics are the survey's synthesis; the listed studies provide mechanism descriptions or empirical contrasts under their respective protocols.}
\label{tab:cross_role_mechanisms}
\centering
\begingroup
\small
\setlength{\tabcolsep}{5pt}
\renewcommand{\arraystretch}{1.10}
\begin{tabularx}{\textwidth}{
  @{}
  >{\raggedright\arraybackslash}p{.35\textwidth}
  >{\raggedright\arraybackslash}p{.30\textwidth}
  >{\raggedright\arraybackslash}X
  @{}
}
\toprule
\TableHeader{Research theme and evidence}
& \TableHeader{Conditional design implication}
& \TableHeader{Proposed diagnostic} \\
\midrule

\textbf{Decisions and completion (RQ1).}
MuSe~\cite{wang2026muse} returns views, GraphChain~\cite{wei2025graphchain} computes results, and LLMCompiler~\cite{kim2024llmcompiler} binds completed outputs. Graph Explorer~\cite{chen2026graphexplorer} separates supported completion from fallback.
& Distinguish missing evidence, interpretation errors, and unmet prerequisites when using feedback. Define completion for the required output.
& Substitute reference specifications and executors separately; record visible choices, outputs, stopping, fallback, failures, and cost. \\
\addlinespace[5pt]

\textbf{Memory and later use (RQ2).}
Relative performance of graph and flat memory varies with returned-value configurations~\cite{hu2026memoryneed}; effects of fact compression and context granularity depend on questions and readers~\cite{wu2024longmemeval}. Action-oriented evidence comes from AriGraph~\cite{anokhin2025arigraph}; team-memory evidence comes from G-Memory~\cite{zhang2025gmemory}, under separate protocols.
& Match retained sources, recovered detail, and update timing to the later task. Separate current applicability, history, and deletion.
& Advance updates chronologically; vary context under declared budgets. Inspect stored and retrieved state, saved arguments, and team use separately. \\
\addlinespace[5pt]

\textbf{Dependency preservation (RQ3).}
CWQ node prediction and path QA diverge under sparsification in GNN-RAG~\cite{mavromatis2025gnnrag}; sparse routes reduce the influence of both harmful and useful supplied information in EIB-LEARNER's~\cite{shen2025topologypropagation} topology-specific tests. GraSP~\cite{xia2026grasp} checks invocation conditions and repairs dependencies.
& Preserve usable support, invocation requirements, and message routes; revalidate assumptions affected by repair.
& Track paths, readiness, effects, and messages. Distinguish fixed-message replay from interactive reruns; account for full-workload costs. \\

\bottomrule
\end{tabularx}
\endgroup
\end{table*}

\section{Cross-Role Synthesis: Decisions, Memory, and Dependencies}
\label{sec:cross_cutting}

Sections~\ref{sec:reasoning_objects}--\ref{sec:agentic_graphs} review mechanisms by graph function. Here we compare their downstream effects through three questions: how observations and computation guide decisions (RQ1), how retained information supports later tasks (RQ2), and which dependencies need to be preserved (RQ3). These questions span roles: a retrieved fact can become an action argument, and its correction may need to reach another agent before execution. Table~\ref{tab:cross_role_mechanisms} separates the reported evidence from our design implications and proposed diagnostics.

\subsection{Observation, Computation, and Completion}
\label{subsec:mechanism_coupling}
\label{subsec:graph_interfaces}

Observation and delegation allocate computation differently. MuSe~\cite{wang2026muse} returns a local view for model reasoning, GraphChain~\cite{wei2025graphchain} performs a requested graph operation, and LLMCompiler~\cite{kim2024llmcompiler} binds completed outputs to dependent calls. View extraction leaves target inference to the model. Delegation requires correct specification and argument binding before execution and result interpretation afterward; dispatch additionally depends on readiness. An executor can correctly execute a specification that does not match the intended task. These responsibilities motivate separate tests of specifications, execution, and subsequent use.

Feedback should address the source of unfinished work. KG exploration acquires support~\cite{sun2024think}, while query execution combines retrieved facts through operators such as joins and counting~\cite{luo2024chatkbqa}. Plan-on-Graph~\cite{chen2024planongraph} extends search or backtracks when support is insufficient; VoG~\cite{zhao2026vog} revises plans when predicted observations conflict with retrieved facts. Interleaving Retrieval with Chain-of-Thought (IRCoT)~\cite{trivedi2023interleaving} provides a non-graph counterpart in which reasoning redirects text retrieval. The relevant distinction is whether feedback changes evidence acquisition or the interpretation that governs its use.

Completion depends on the required output. ARK~\cite{polonuer2026autonomous} returns a ranked list when it selects \texttt{FINISH} or reaches a trajectory limit, while Graph Explorer~\cite{chen2026graphexplorer} distinguishes valid graph-supported completion from best-effort answering. A correct fallback answer may leave the supported-completion criterion unmet.

The value of additional graph access is task-dependent. ARK's sampled-test ablation shows larger retrieval losses from removing neighborhood access on STaRK's~\cite{wu2024stark} academic MAG (Microsoft Academic Graph) and biomedical PRIME subsets than on its product-oriented AMAZON subset~\cite{polonuer2026autonomous}. KARPA~\cite{fang2025karpa} and CoR~\cite{liu2026chainrelations} report fewer calls than their baselines, whereas MuSe's~\cite{wang2026muse} round-limit ablation concerns training. These results assess different forms of interaction control.

For RQ1, the reviewed mechanisms use three types of decision input: available evidence, computed results, and execution status. Their usefulness depends on the remaining task, its completion rule, and interaction cost. The reviewed comparisons do not establish a general policy for choosing observations or computations and deciding when to stop (Section~\ref{subsec:future_conditional_utility}).

\subsection{Memory Representations and Later Use}
\label{subsec:retained_state_synthesis}

Search representations and answering content serve separate purposes. Hu et al.'s~\cite{hu2026memoryneed} selected graph and flat configurations reverse their relative performance across returned-value settings on filtered question subsets with unequal source detail. The LongMemEval$_M$ analysis~\cite{wu2024longmemeval} reports lower overall QA performance with compact facts, despite gains in multi-session reasoning; granularity effects also depend on the reader. These results favor separate choices about stored sources, on-demand recovery, and immediate context, without isolating a topology-only effect.

Relations support different stages of evidence use. Entity--session associations~\cite{hu2026memoryneed} support recovery of exchanges, SGMem~\cite{wu2025sgmem} uses sentence and parent-chunk links to recover context, and RAPTOR's~\cite{sarthi2024raptor} hierarchy organizes summary construction. Hu et al.'s no-expansion configuration still uses alignment and source mapping. Graph contribution can therefore arise in preparation or recovery without online neighbor traversal. Such associations, however, need not record how each derived assertion depends on its sources.

Maintenance determines both applicability and availability. Note evolution~\cite{xu2025amem}, conflict invalidation with history~\cite{chhikara2025mem0}, temporal validity~\cite{rasmussen2025zep}, and deferred consolidation~\cite{fang2025lightmem} expose different states to later requests. Current and historical questions require the corresponding valid assertions; requested deletion concerns retention itself. Measurements taken only before and after consolidation do not characterize what a request can access during an incomplete update. Update timing is therefore part of the task setting, alongside storage and context budgets.

Planning and collaboration further change the required content. AriGraph~\cite{anokhin2025arigraph} supplies environmental state and episodes for action, while G-Memory~\cite{zhang2025gmemory} supplies role-specific insights and interaction traces to later teams. Their separate evaluations extend the evidence beyond dialogue QA, but do not establish whether correcting a stored assertion also changes a saved action argument or shared guidance. Testing this cross-role composition requires tracing corrections from stored assertions into the plans and guidance that use them.

For RQ2, the evidence favors separating stored content, retrieval units, and answering context. Their design depends on the later task, reader, budget, and retention constraints. Update timing determines which state is available; whether a memory correction reaches saved plans or team decisions requires joint testing.

\subsection{Dependencies and Downstream Behavior}
\label{subsec:selective_activation_search}
\label{subsec:verification_feedback_repair}

Selection matters through the relations left available downstream. In GNN-RAG's~\cite{mavromatis2025gnnrag} answer-containing candidate-subgraph setting, CWQ sparsification improves node prediction but lowers path-based answering. AgentPrune~\cite{zhang2025agentprune} and AgentDropout~\cite{wang2025agentdropout} reduce communication work in their evaluated settings. EIB-LEARNER's~\cite{shen2025topologypropagation} interventions use topology-specific correctness-conditioned question sets; degree-normalized answer changes indicate that sparse routes suppress the influence of both induced errors and supplied correct answers. The shared concern is usable support, although evidence paths and message routes have different semantics. These study-specific comparisons leave the preferred graph density dependent on the task, available support, and downstream operation.

Execution requires applicable prerequisites as well as available information. ToolNet~\cite{liu2024toolnet} ranks successors, ControlLLM~\cite{liu2023controlllm} identifies resource-compatible compositions, and LLMCompiler~\cite{kim2024llmcompiler} binds predecessor results for dispatch. GraSP~\cite{xia2026grasp} adds checks of represented invocation conditions and bounded repair. If the recorded graph omits a shared-state dependency, affected operations may lie outside the recorded descendants of the modified nodes. This mechanism-based inference motivates validating the assumptions affected by repair, beyond counting edited nodes.

Verification should be selected for the property being assessed. GCR~\cite{luo2025graphconstrained} constrains indexed-path membership, ChatKBQA~\cite{luo2024chatkbqa} computes query denotations, and GraSP~\cite{xia2026grasp} checks represented invocation conditions. Additional checks are needed to assess whether the interpretation matches the task and whether execution achieves the intended effect.

Consider the facility example from the introduction. At time $t$, the operator changes from $O_1$ to $O_2$. Memory records $O_2$ as current and retains $O_1$ for historical queries, but a saved notification plan still targets $O_1$. Retrieval and message delivery can both succeed while the request to notify the current operator fails. This illustrative scenario links fact validity, argument binding, and dispatch: the current notification should reach $O_2$, while a historical query for $O_1$'s period of operation should return $O_1$.

For RQ3, relation selection should preserve the evidence connections required by the task, applicable invocation requirements, and routes for decision-relevant messages. These dependencies matter through the operations that use them. Section~\ref{sec:evaluation} translates the synthesis for all three research questions into comparison designs, process diagnostics, and cost measures.

\begin{table*}[t]
\caption{Benchmarks and their relevance to decisions and completion ($RQ_1$), memory and later use ($RQ_2$), and dependency preservation ($RQ_3$). RQ labels indicate relevance rather than complete coverage of a research question.}
\label{tab:benchmark_landscape}
\centering
\begingroup
\small
\setlength{\tabcolsep}{4.5pt}
\renewcommand{\arraystretch}{1.08}
\begin{tabularx}{\textwidth}{
  @{}
  >{\raggedright\arraybackslash}p{.14\textwidth}
  >{\raggedright\arraybackslash}p{.29\textwidth}
  >{\raggedright\arraybackslash}p{.27\textwidth}
  >{\raggedright\arraybackslash}X
  @{}
}
\toprule
\TableHeader{Evaluation target}
& \TableHeader{Representative resources}
& \TableHeader{Observable evidence}
& \TableHeader{Interpretation boundary} \\
\midrule

\textbf{Graph computation}\newline $RQ_1$
& GraphArena~\cite{tang2025grapharena}; MAGMA~\cite{taylor2024magma}; GraphEval36K~\cite{wu2025grapheval36k}; GraphOmni~\cite{xu2026graphomni}; GraphGym~\cite{wang2026unified}.
& Solution validity and optimality; supplied-state transitions; program tests; representation and loading sensitivity.
& Correct conditional transitions and passing test cases do not establish reliable autonomous execution. \\
\addlinespace[4pt]

\textbf{KG interpretation}\newline $RQ_1$, $RQ_3$
& WebQSP~\cite{yih2016webqsp}; CWQ~\cite{talmor2018complexwebquestions}; KQA Pro~\cite{cao2022kqapro}; GrailQAbility~\cite{patidar2023grailqability}.
& Semantic parses, compositional questions, explicit programs, and missing-data or schema conditions.
& Annotation types differ; a correct denotation can arise from an incorrect interpretation. \\
\addlinespace[4pt]

\textbf{Evidence access}\newline $RQ_1$, $RQ_3$
& HotpotQA~\cite{yang2018hotpotqa}; 2WikiMultiHopQA~\cite{ho2020twowiki}; MuSiQue~\cite{trivedi2022musique}; STaRK~\cite{wu2024stark}; Xiang et al.'s evaluation~\cite{xiang2026whentousegraphs}; GraphRAG-Bench~\cite{xiao2026graphragbench}.
& Supporting sentences, evidence triples, connected question compositions, retrieval ranks, and stage-level outcomes.
& Retrieval relevance, sufficient support, and faithful citation require separate assessment. \\
\addlinespace[4pt]

\textbf{Dialogue memory}\newline $RQ_2$
& LoCoMo~\cite{maharana2024locomo}; LongMemEval~\cite{wu2024longmemeval}; HaluMem~\cite{chen2025halumem}.
& Long-history QA, temporal and update-dependent questions, and extraction or update assessments.
& Answer-based and retrieval-mediated tests cover different aspects of retained state, not necessarily the entire store. \\
\addlinespace[4pt]

\textbf{Interactive execution}\newline $RQ_1$, $RQ_3$
& TextWorld~\cite{cote2018textworld}; ALFWorld~\cite{shridhar2021alfworld}; ScienceWorld~\cite{wang2022scienceworld}; WebShop~\cite{yao2022webshop}; AgentBench~\cite{liu2024agentbench}; ToolBench~\cite{qin2024toolllm}.
& Environment progress, action effects, and tool trajectories; ToolEval~\cite{qin2024toolllm} provides judged outcomes on ToolBench.
& Invocation success, task success, and evaluator acceptance differ. Cross-task retention requires an explicit protocol. \\
\addlinespace[4pt]

\textbf{General outcomes}\newline $RQ_1$, $RQ_3$
& HumanEval~\cite{chen2021humaneval}; MMLU~\cite{hendrycks2021mmlu}.
& Unit-test correctness and multiple-choice accuracy, including outcomes used in workflow and communication studies.
& These annotations alone do not test persistent updates or naturally discovered corrections. \\

\bottomrule
\end{tabularx}
\endgroup
\end{table*}

\section{Evaluation of Graph Contributions}
\label{sec:evaluation}

Evaluation connects the mechanisms reviewed in Sections~\ref{sec:reasoning_objects}--\ref{sec:agentic_graphs} and the cross-role synthesis in Section~\ref{sec:cross_cutting} to testable claims. \emph{Outcome metrics} assess answer quality and task success; \emph{process evidence} records relation use, state changes, execution readiness, and message delivery; and \emph{system metrics} quantify resource use and costs.

Benchmark annotations and evaluation protocols determine which aspects of these claims can be observed. Graph-computation and KGQA tasks assess computational correctness and interpretation; evidence and memory tasks examine support and temporal use; and interactive environments expose action consequences. Table~\ref{tab:benchmark_landscape} summarizes representative resources, their relevance to RQ1--RQ3, and the limits of their observable evidence. Task scores alone cannot establish claims about memory updates, supported completion, or information delivery~\cite{yehudai2026evaluation}. We therefore consider three complementary aspects of evaluation: comparison designs and experimental controls, process diagnostics, and system costs and reporting requirements.

\subsection{Comparison Designs and Experimental Controls}
\label{subsec:eval_framework}
\label{subsec:attribution_targets}

The comparison target determines what should be held fixed. A \emph{whole-pipeline comparison} evaluates construction, access, and downstream reasoning or execution together. It uses common raw inputs and tasks under declared resource conditions, while allowing intermediate summaries, retrieved evidence, memories, and trajectories to differ. A \emph{conditional online comparison} fixes upstream inputs, such as an index for retrieval, a candidate set for reranking, or selected evidence for testing its arrangement. A \emph{component intervention} replaces or disables one operation, measures the resulting behavior, and specifies whether downstream components may adapt. The first two designs define comparison scope; component interventions can be used within either scope.

RAPTOR's~\cite{sarthi2024raptor} tree traversal and collapsed retrieval operate over a constructed summary hierarchy. Comparing them on the same index targets online access; evaluating the complete approach also includes summary construction and its cost. Holding retrieved content fixed removes differences in selected evidence from the comparison; holding post-update memory fixed similarly removes differences in retained state. These tests exclude effects mediated by those differences, although the costs of selection or maintenance may still vary. Equal top-$k$, token, or call limits constrain budgets without equalizing information, capabilities, or realized costs.

Structural interventions require task-consistent labels. For graph-computation tasks, changing an input edge may change the correct answer. Removing evidence from a knowledge graph may instead change what the system can establish without changing the underlying fact. Removing a dependency edge from a controller does not remove the environment's actual prerequisites. Redundancy or adaptation can mask an intervention's effect, so unchanged outcomes alone do not demonstrate that a relation was unused. Reports should specify the altered relations, valid answers, available alternatives, and permitted adaptation.

\subsection{Process Diagnostics}
\label{subsec:structural_validity_robustness}
\label{subsec:evidence_adequacy_utilization}

\noindent\textbf{Decisions and completion (RQ1).}
Records should identify available candidates, selected operations, arguments, returned results, and stopping reasons. Where reference specifications and executors are available, running a reference specification with the deployed executor tests execution against an expected result. Running a model-produced specification with a reference executor establishes the result of that specification; a separate semantic check is still needed to determine whether it expresses the intended task. These substitutions diagnose different interfaces and complement end-to-end evaluation.

Completion rules also define the measured outcome. Graph Explorer's~\cite{chen2026graphexplorer} finish-or-fail protocol requires a correct answer through a valid \texttt{Finish} within budget and at least one KG call; best-effort accuracy and post-hoc visibility checks assess separate properties. Path conformance in GCR~\cite{luo2025graphconstrained} and query denotation in ChatKBQA~\cite{luo2024chatkbqa} should likewise be assessed alongside semantic alignment. Citation evaluation further distinguishes full, partial, and absent support~\cite{zhang2024citationfaithfulness}. SIRG~\cite{hu2026sirg} constructs semantic-level attribution graphs to detect RAG faithfulness hallucinations, diagnosing internal dependencies rather than validating source-evidence paths.

\noindent\textbf{Memory and later use (RQ2).}
Interactions should advance chronologically from a common initial history, with resets and update-completion times reported and future information excluded. Memory representations and updated states may differ across systems when those mechanisms are under evaluation. Tests should separately assess current-state answers, historical answers, correction handling, and requested deletion. HaluMem~\cite{chen2025halumem} retrieves ten relevant memories per annotated update, making its update assessment retrieval-mediated; unavailable extraction metrics for Zep reflect the missing interface in that protocol. Storage inspection, retrieved content, and later answers or actions therefore provide complementary evidence. For the example in Section~\ref{subsec:selective_activation_search}, tests should trace whether the updated operator identity reaches the saved recipient argument and the scheduler's notification decision.

\noindent\textbf{Dependency preservation (RQ3).}
For evidence use, diagnostics should trace the relations available to the answering step. To isolate dependency handling in execution, dispatch comparisons should hold skills and the initial environment state fixed. Repair tests should record revised plans, checker verdicts, the work revalidated, and subsequent effects. ToolEval's~\cite{qin2024toolllm} pass criterion can accept giving up after exhaustive, unproductive API attempts or refusing an unsolvable instruction. Judged acceptance and fulfillment of the requested action consequently require separate reporting.

To isolate routing effects, comparisons should hold participants, capabilities, and aggregation rules fixed. Fixed-message replay controls the content supplied to agents; interactive reruns also capture how subsequent messages and decisions change. EIB-LEARNER~\cite{shen2025topologypropagation} applies information-injection interventions to MMLU and assesses their influence through answer changes. Interpretation of these results should account for the topology-specific, correctness-conditioned question sets. Assessing correction discovery also requires recording how the supporting evidence was obtained. These tests operationalize the distinctions in Table~\ref{tab:cross_role_mechanisms}.

\subsection{System Costs and Reporting Requirements}
\label{subsec:system_evaluation}
\label{subsec:evaluation_protocol}

Full-system accounting covers relevant extraction, indexing, summarization, training or workflow search, online operations, verification, retries, failed runs, and memory updates. Query volume, corpus turnover, and update frequency determine amortization; deferred consolidation remains maintenance work. Aggregate compute usage, wall-clock time, critical-path latency, and tail latency capture different aspects of system performance. Accounting boundaries also differ across studies: Hu et al.'s~\cite{hu2026memoryneed} construction and retrieval costs exclude answering, whereas Mem0's~\cite{chhikara2025mem0} response latency includes generation.

Quality--cost comparisons should report the complete configuration, including content preparation, corpus and task conditions, query routing, and extraction or generation models~\cite{chen2026isgraphragneeded,xiang2026whentousegraphs,dong2026usegraph,han2025rag}. With Qwen2.5-7B-Instruct~\cite{yang2024qwen25} on HotpotQA, Fan et al.~\cite{fan2026ragsearch} report Contain Exact Match scores of 33.76 for dense retrieval and 32.73 for the backend labeled GraphRAG under Search-o1~\cite{li2025searcho1}. Under GraphSearch~\cite{yang2025graphsearch}, dense retrieval scores 38.22 and the same GraphRAG backend scores 42.25. This reversal concerns the specified backend, metric, and configurations; it motivates reporting the search workflow, retrieval backend, and workload costs together.

Reports should identify the operation and paper version, comparison target, tool capabilities, temporal horizon, and cost-accounting boundary. Overall task-success rates should include all evaluated episodes; conditional scores should state the conditioning event and sample size. Failures, timeouts, abstentions, and unresolved checks should be reported. Judge-based evaluation~\cite{zheng2023llmjudge} should specify the rubric, judge model, response ordering, and invalid-result handling, because reported scores can depend on position and verbosity biases. Queries sharing retained memory should be analyzed as dependent sequences.

\section{Open Challenges and Future Directions}
\label{sec:challenges}

Building on the cross-role synthesis in Section~\ref{sec:cross_cutting} and the evaluation framework in Section~\ref{sec:evaluation}, we identify three directions for future research: adaptive observation, computation, and stopping (RQ1); reliable memory maintenance and subsequent use (RQ2); and dependency-aware recovery and correction propagation (RQ3).

\subsection{Adaptive Computation and Stopping}
\label{subsec:future_conditional_utility}
\label{subsec:future_structural_procedures}

\noindent\textbf{Operation value and diagnostic feedback.}
The value of another operation depends on the information or computation still needed. IRCoT~\cite{trivedi2023interleaving} redirects text retrieval, Plan-on-Graph~\cite{chen2024planongraph} tracks sub-objectives, and Execution as Verification~\cite{zhang2026executionverification} uses execution feedback to revise queries. A broader policy would distinguish missing evidence, mistaken interpretation, and execution failure before choosing another operation. Training could compare alternative continuations from a common task state, including their acquisition and verification costs. Because the value of an observation can depend on later operations, these tests should allow subsequent decisions to adapt.

\noindent\textbf{Transfer of structural procedures.}
G1~\cite{guo2025g1} studies transfer across graph tasks and encodings; G-Reasoner~\cite{luo2026greasoner} addresses retrieval over heterogeneous graph-structured knowledge. These settings motivate separating representation invariance from adaptation to changed semantics or tool preconditions. One direction is to combine verifiable state transitions with feedback-driven tool trajectories and explicit operation contracts. Experience reuse, as in Reflexion~\cite{shinn2023reflexion}, could help revise a procedure when its assumptions fail. Evaluation should report both transfer performance and the interaction cost of adaptation.

\noindent\textbf{Calibrated stopping.}
A stopping policy needs to assess both current support and the potential value of further access. Fully specified graph tasks may admit checkable completion conditions; unsuccessful search in an incomplete knowledge graph may leave answerability unresolved. Policies should distinguish supported completion, abstention for insufficient evidence, and budget exhaustion. Confidence alone may be inadequate when tool reliability or unverified dependencies affect the decision. Evaluation should report premature termination, fallback answers, failed runs, and resource use, with task consequences reflected in the cost of stopping or continuing.

\subsection{Memory Maintenance for Later Decisions}
\label{subsec:future_long_horizon}

\noindent\textbf{Selective retention and source recovery.}
Retention should preserve the information needed for later tasks within storage and context budgets. LongMemEval~\cite{wu2024longmemeval}, Thought-Action Graph Reasoning~\cite{qi2026thought}, and AriGraph~\cite{anokhin2025arigraph} illustrate different needs for factual, procedural, and environmental information. A research direction is to learn which relational and temporal qualifiers should survive compression and when to recover omitted source detail. Recovery depends on the source remaining available and eligible for retention. Procedural reuse also requires checking current tool capabilities and invocation conditions.

\noindent\textbf{Consistency during maintenance.}
Existing methods revise notes~\cite{xu2025amem}, invalidate conflicting relations~\cite{chhikara2025mem0}, represent temporal validity~\cite{rasmussen2025zep}, or defer consolidation~\cite{fang2025lightmem}. A request arriving during maintenance may read a revised fact alongside an older summary or cached plan. Versioned records and pending-update states could identify affected operations, allowing them to refresh their inputs or wait while unrelated work continues. The time for which a fact is valid should be distinguished from the time at which its update completes. LongMemEval~\cite{wu2024longmemeval} and HaluMem~\cite{chen2025halumem} offer starting points, but evaluation of interrupted updates also needs decisions made before maintenance finishes.

\noindent\textbf{Provenance for derived state.}
Source references identify related evidence but may not show how a generated summary or plan depends on it. Provenance semirings distinguish alternative derivations from jointly used inputs for positive relational queries~\cite{green2007provenance}. Generated content can additionally depend on temporal assumptions and model knowledge. Partial dependency records could distinguish verified support, heuristic associations, and unknown dependencies, enabling selective revalidation without claiming a complete derivation. Historical access and deletion also require different handling: invalidation can retain past support, whereas deletion needs to address source, derived, and cached records within the requested scope. Evaluation should weigh recording and revalidation costs against the consequences of using stale information downstream.

\subsection{Dependency-Aware Recovery and Information Delivery}
\label{subsec:future_verifiable_execution}
\label{subsec:future_coordination_credit}

\noindent\textbf{Recovery with incomplete dependency records.}
Selective repair depends on identifying the work affected by a change. GraSP~\cite{xia2026grasp}, FLOW~\cite{niu2025flow}, and HARNESSFIX~\cite{chen2026harnessfix} address invocation repair, workflow revision, and harness diagnosis, respectively. A database precedent is the Delete and Rederive (DRed) algorithm~\cite{gupta1993maintaining}, which overdeletes potentially affected tuples and rederives those with surviving support under known rules. Reads, writes, and external effects can create dependencies not recorded in an execution plan. A repair mechanism could combine declared dependencies with observed accesses and expand the scope of revalidation when effects are uncertain. Evaluation should distinguish patch size, affected work revisited, and valid work preserved.

\noindent\textbf{Verification of external effects.}
State can change between a precondition check and a tool call. Version-aware argument binding and checks near dispatch may help, but their guarantees depend on the external interface. Recovery should distinguish observed success, observed failure, and unresolved outcomes: a timed-out call may already have changed the environment, so retrying can duplicate an effect. Execution graphs could record observable effects, retry conditions, and available compensation. Tests should compare reductions in execution errors with the added verification cost and delay. When rollback is unavailable and compensation is possible, the compensating action should be assessed with its own prerequisites and costs and should respect task requirements and authorization constraints.

\noindent\textbf{Delivery and use of corrections.}
Correction requires evidence to reach a component that can act on it. AMAS~\cite{leong2025amas} and GTD~\cite{jiang2026gtd} select task-specific topologies, while EvoHyper~\cite{zhang2026evohyper} studies group structure and shared memory. EIB-LEARNER~\cite{shen2025topologypropagation} measures the effects of supplied errors and correct answers. A broader target is the path from discovering a correction to using it before a dependent decision. Pending operations could guide recipient selection, while messages identify the revised assertion, its valid time, and the assumption replaced. Evaluation should distinguish discovery, delivery, and use. Fixed-message replay controls the supplied content; interactive reruns also capture how newly obtained evidence and subsequent decisions change.

Integrity checks may also obstruct valid corrections. AiTM~\cite{he2025aitm} studies message attacks, and GUARDIAN~\cite{zhou2025guardian} models temporal interaction graphs for anomaly detection. Routing policies should limit unreliable propagation while allowing justified disagreement. Provenance can help distinguish independent support from repeated transmission, but source identity alone does not establish truth. Evaluation should report both acceptance of incorrect proposed corrections and rejection of valid ones, including cases with changed participants or tools.

Across these directions, evaluation should track which decisions are affected by new information, whether the affected work is corrected, and at what cost. Comparisons with flat-memory, fixed-workflow, and fixed-routing baselines can clarify the benefits and costs of relational organization and runtime adaptation, provided other relevant differences are controlled.

\section{Conclusion}
\label{sec:conclusion}

This survey organizes graph-mediated LLM systems by graph function and computational responsibility. Operation-level comparisons distinguish observation from delegated computation, searchable representations from answering context, and planned dependencies from runtime enforcement. Cross-role analysis traces how these mechanisms affect later decisions, memory use, execution, and communication. The evaluation framework links these claims to benchmarks, comparison designs, process evidence, and workload costs. The remaining challenges concern adaptive computation and stopping, reliable memory maintenance, and recovery and correction propagation when dependency information is incomplete. These distinctions provide a structured basis for designing and evaluating graph-mediated LLM systems.

\bibliographystyle{IEEEtran}
\bibliography{references}

\makeatletter
\renewcommand{\@IEEEBIOskipN}{2.5\baselineskip}
\makeatother

\begin{IEEEbiography}[{\includegraphics[width=1in,height=1.25in,clip,keepaspectratio]{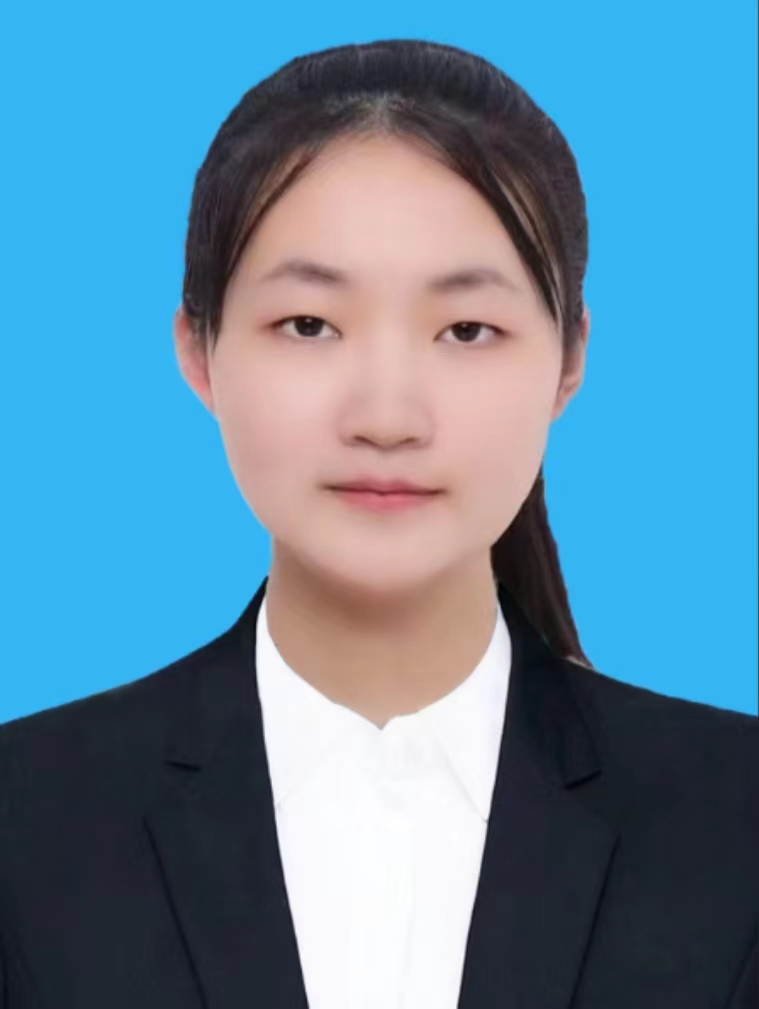}}]{Haihui Liu} (Student Member, IEEE) received the M.S. degree from Shandong Normal University, China, in 2024. She is currently working toward the Ph.D. degree at Beijing Normal University. Her research interests include edge intelligence, recommender systems, and large language models. 
\end{IEEEbiography}

\begin{IEEEbiography}[{\includegraphics[width=1in,height=1.25in,clip,keepaspectratio]{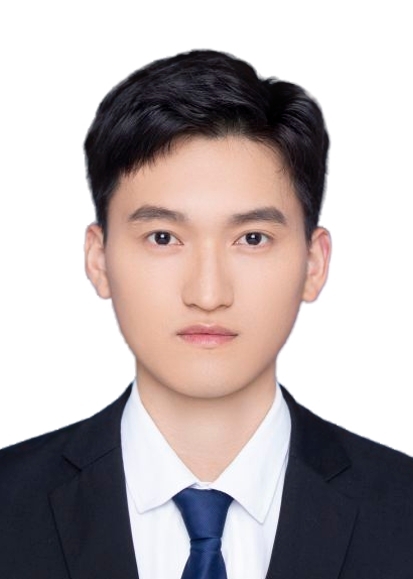}}]{Yang Chen} received the B.Sc. degree in information and computing science from Nanjing Forestry University, Nanjing, China, in 2023, and the M.Eng. degree in computer science and technology from South-Central Minzu University, Wuhan, China, in 2026. He is currently working toward the Ph.D. degree with the Faculty of Arts and Sciences, Beijing Normal University, Zhuhai, China. His research interests include graph-based agents and knowledge graph question answering.
\end{IEEEbiography}

\begin{IEEEbiography}[{\includegraphics[width=1in,height=1.25in,clip,keepaspectratio]{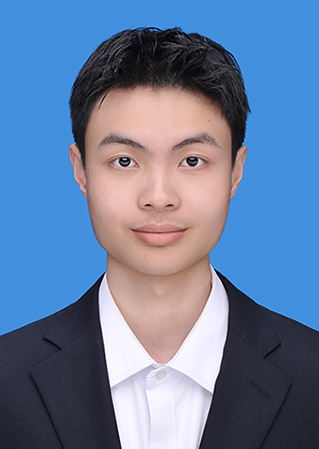}}]{Zice Qiu} is currently working toward the bachelor's degree in remote sensing science and technology at the School of Geospatial Engineering and Science, Sun Yat-sen University, Zhuhai, China. His research interests include large language models, multimodal learning, and AI-generated content.
\end{IEEEbiography}

\begin{IEEEbiography}[{\includegraphics[width=1in,height=1.25in,clip,keepaspectratio]{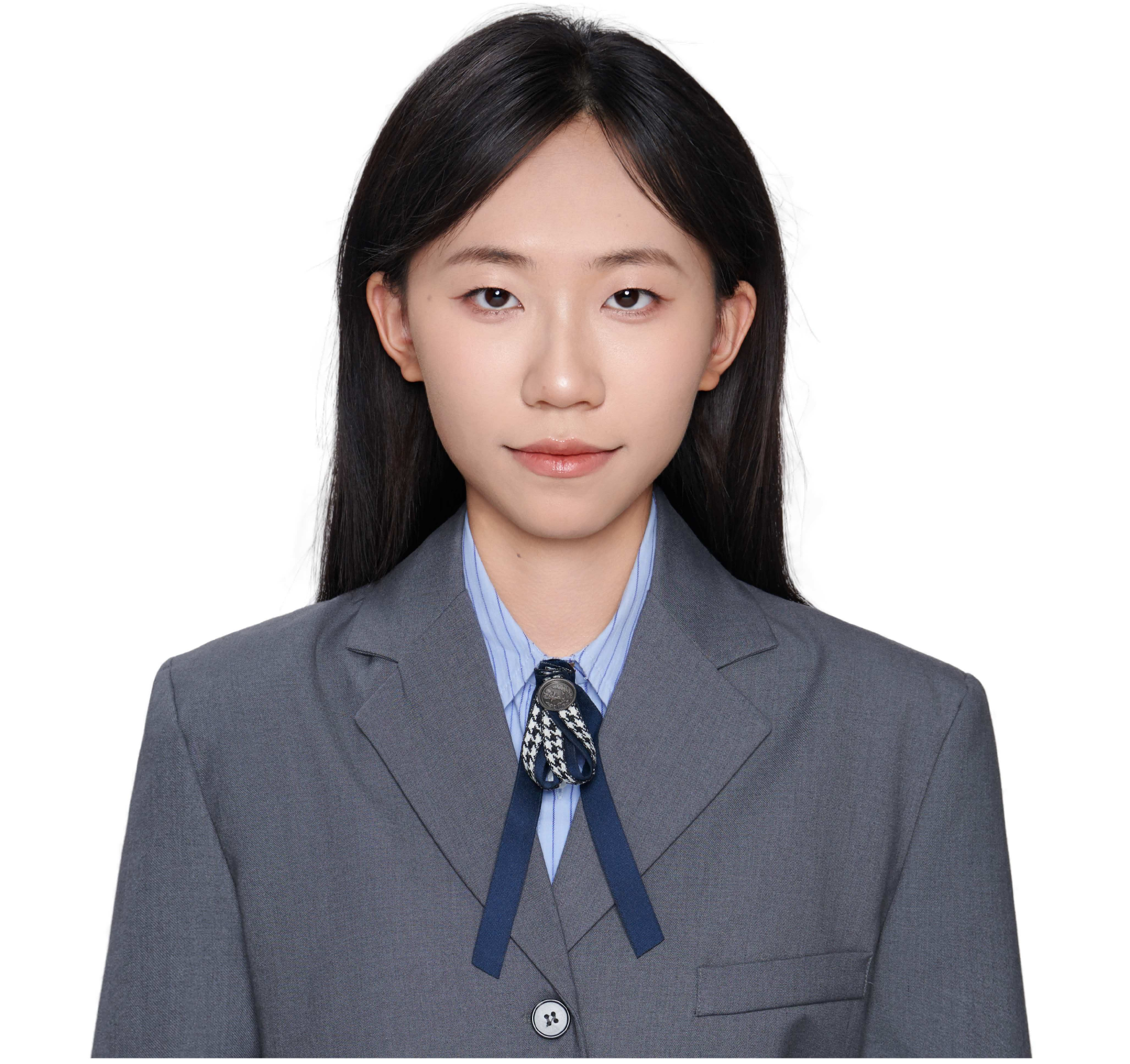}}]{Yun Zhang} is currently pursuing her undergraduate degree at the Faculty of Science and Technology, BEIJING NORMAL‑HONG KONG BAPTIST UNIVERSITY. Her research interests include data analysis and artificial intelligence.
\end{IEEEbiography}

\begin{IEEEbiography}[{\includegraphics[width=1in,height=1.25in,clip,keepaspectratio]{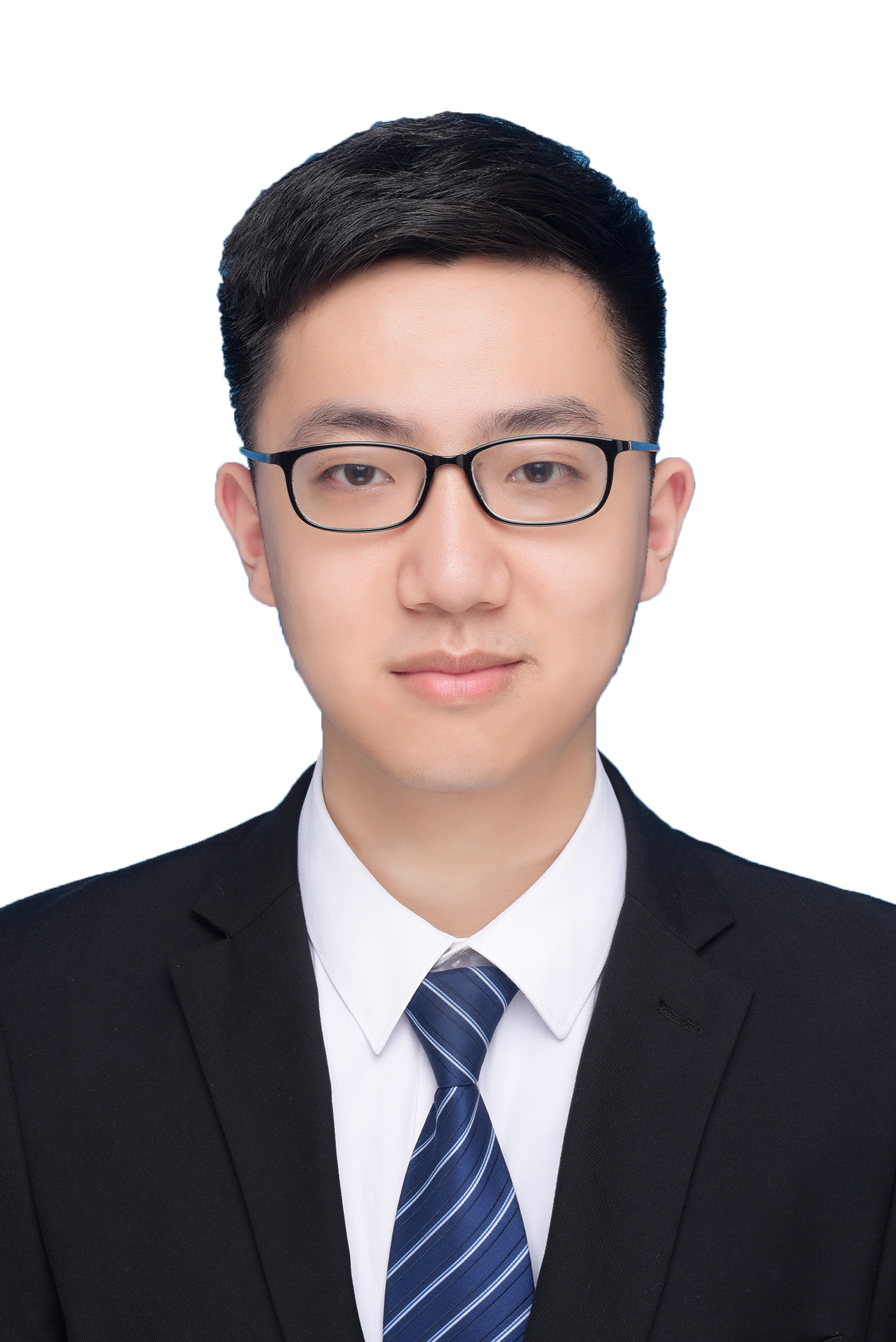}}]{Jialun Zhong} received the M.Sc. degree from Huazhong University of Science and Technology, China, in 2023. He is currently working toward a Ph.D. degree at the School of Intelligence Science and Technology, Peking University, China. His research interests include knowledge graphs and large language model reasoning.
\end{IEEEbiography}

\begin{IEEEbiography}[{\includegraphics[width=1in,height=1.25in,clip,keepaspectratio]{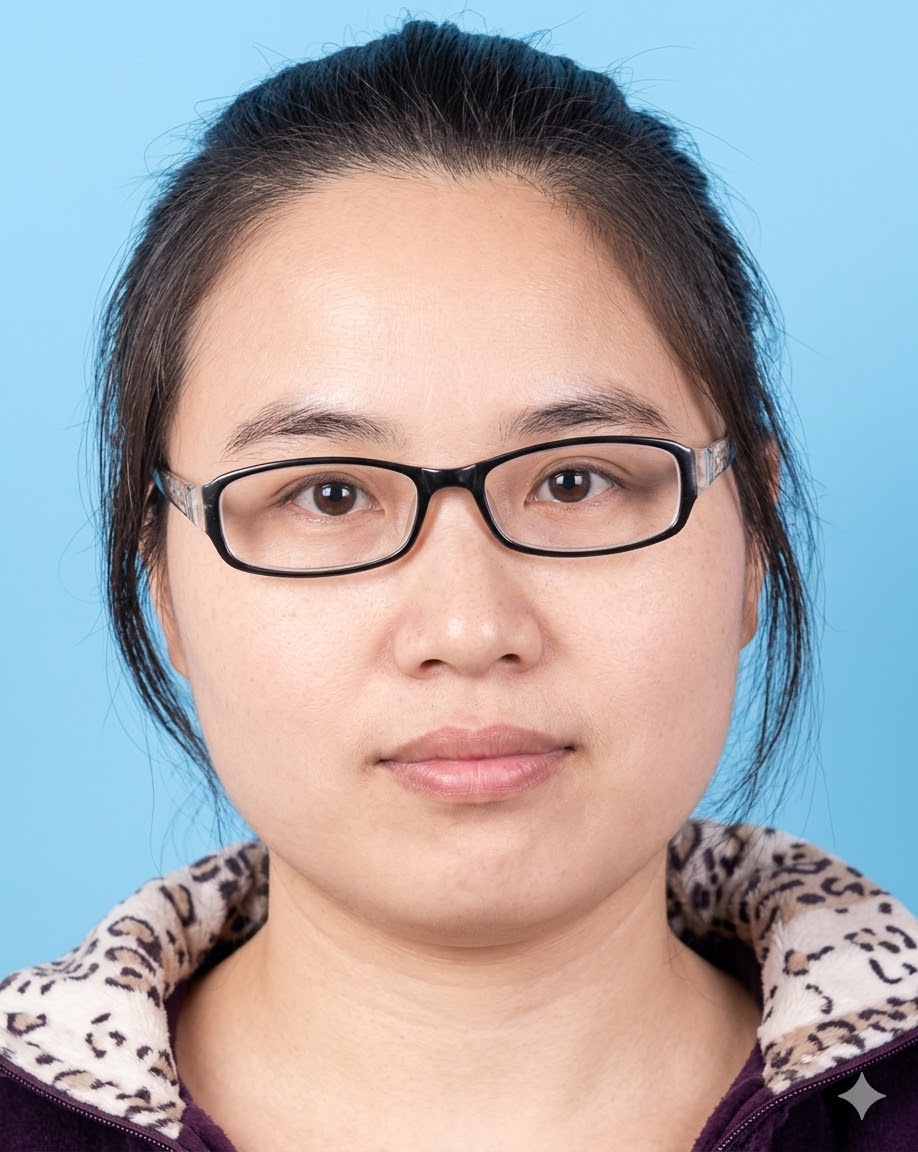}}]{Wenhua Wang} (Member, IEEE) received the Ph.D. degree in computer science from the University of Technology Sydney in 2022. She is currently a lecturer with the Institute of Artificial Intelligence and Future Network, Beijing Normal University, China. Her research interests include edge computing, federated learning, and mobile computing.
\end{IEEEbiography}

\begin{IEEEbiography}[{\includegraphics[width=1in,height=1.25in,clip,keepaspectratio]{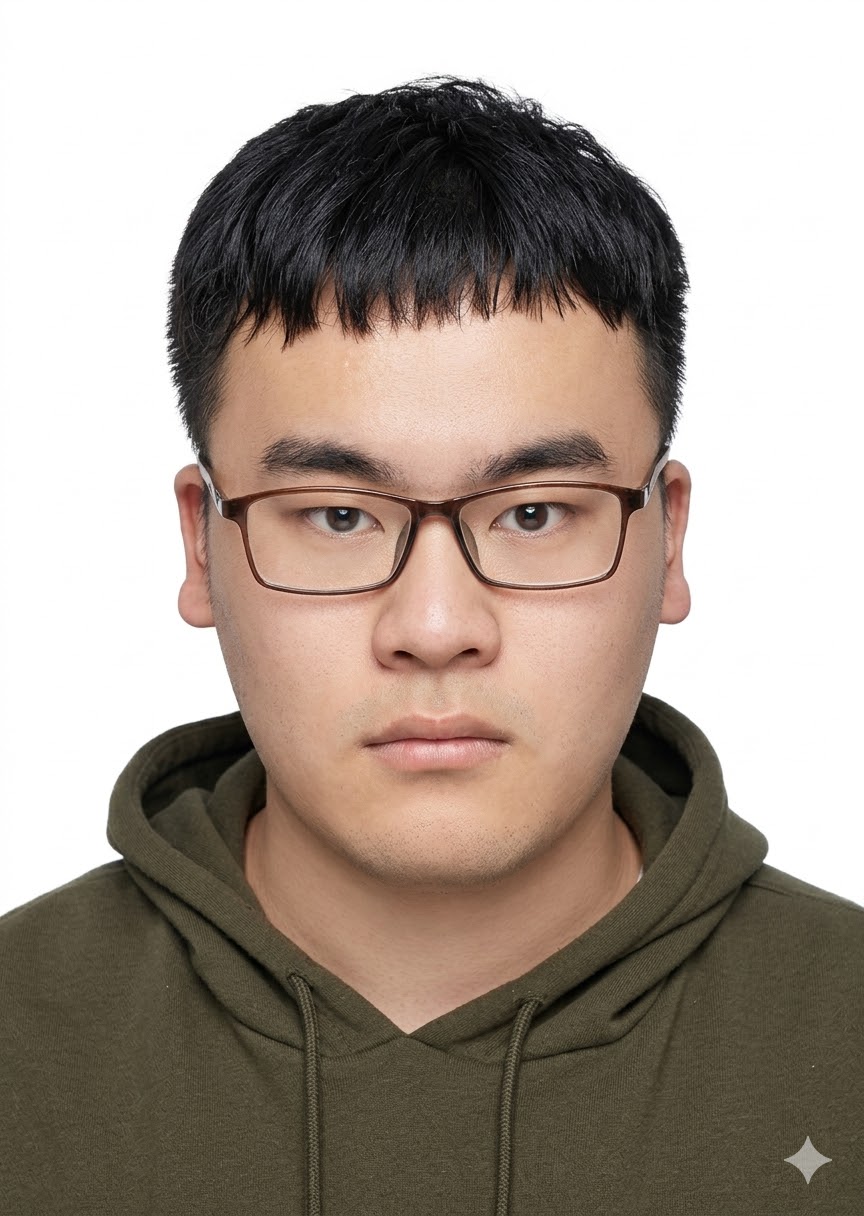}}]{Yanzeng Li} is currently a lecturer at the Institute of Artificial Intelligence and Future Networks, Beijing Normal University, China. He received his Ph.D. degree from Peking University in 2025. His research focuses on knowledge graphs and large language models, and he serves as a reviewer and area chair for numerous international journals and conferences in artificial intelligence and natural language processing. 
\end{IEEEbiography}

\begin{IEEEbiography}[{\includegraphics[width=1in,height=1.25in,clip,keepaspectratio]{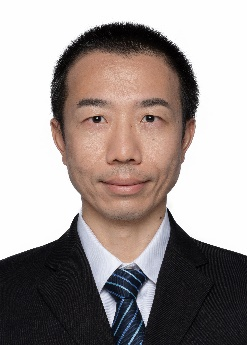}}]{Tian Wang} (Senior Member, IEEE) received the B.Sc. and M.Sc. degrees in computer science from Central South University in 2004 and 2007, respectively, and the Ph.D. degree in computer science from City University of Hong Kong in 2011. He is currently a professor with the Institute of Artificial Intelligence and Future Networks, Beijing Normal University. He has authored or coauthored more than 200 papers in high-level journals and conferences. He has more than 17,000 citations according to Google Scholar and also has 30 patents. His h-index is 78. His research interests include the Internet of Things, edge computing, and mobile computing.
\end{IEEEbiography}

\end{document}